\documentclass[letterpaper]{article} % DO NOT CHANGE THIS
\usepackage{aaai24}  % DO NOT CHANGE THIS
\usepackage{times}  % DO NOT CHANGE THIS
\usepackage{helvet}  % DO NOT CHANGE THIS
\usepackage{courier}  % DO NOT CHANGE THIS
\usepackage[hyphens]{url}  % DO NOT CHANGE THIS
\usepackage{graphicx} % DO NOT CHANGE THIS
\usepackage{natbib}  % DO NOT CHANGE THIS AND DO NOT ADD ANY OPTIONS TO IT
\usepackage{caption} % DO NOT CHANGE THIS AND DO NOT ADD ANY OPTIONS TO IT
\usepackage{booktabs}       % professional-quality tables
\usepackage{multirow}       % table
\usepackage{algorithm}
\usepackage{algorithmic}
\usepackage{amsmath,amsfonts,bm}

\usepackage{subcaption}

\DeclareMathOperator*{\argmin}{arg\,min}

\usepackage[nameinlink,capitalise]{cleveref}

\usepackage{newfloat}
\usepackage{listings}
\DeclareCaptionStyle{ruled}{labelfont=normalfont,labelsep=colon,strut=off} % DO NOT CHANGE THIS
\floatstyle{ruled}
\newfloat{listing}{tb}{lst}{}
\floatname{listing}{Listing}
\newcommand{\ourmethod}{Equity-Transformer}

\usepackage{xcolor}

\newcommand{\changed}[1]{{\color{black}#1}}
\newcommand{\recheck}[1]{{\color{black}#1}}

\title{\ourmethod: Solving NP-hard Min-Max Routing Problems \\as Sequential Generation with Equity Context}
\author{
    Jiwoo Son\equalcontrib\textsuperscript{\rm 1,\rm 2},
    Minsu Kim\equalcontrib\textsuperscript{\rm 1},
    Sanghyeok Choi\textsuperscript{\rm 1},
    Hyeonah Kim\textsuperscript{\rm 1},
    Jinkyoo Park\textsuperscript{\rm 1,\rm 2}
}

\affiliations{
    \textsuperscript{\rm 1}Korea Advanced Institute of Science and Technology (KAIST)\\
    \textsuperscript{\rm 2}Omelet
    \{sonleave25, min-su, sanghyeok.choi, hyeonah\_kim, jinkyoo.park\}@kaist.ac.kr
}

\usepackage{bibentry}
\begin{document}

\maketitle

\begin{abstract}
% Min-max routing problems aim to minimize the maximum tour length among agents as they collaboratively visit all cities, i.e., the completion time. These problems include impactful real-world applications but are known as NP-hard. Existing methods are facing challenges, particularly in large-scale problems that require the coordination of numerous agents to cover thousands of cities. This paper proposes a new deep-learning framework to solve large-scale min-max routing problems. We model the simultaneous decision-making of multiple agents as a sequential generation process, allowing the utilization of scalable deep-learning models for sequential decision-making. In the sequentially approximated problem, we propose a scalable contextual Transformer model, \ourmethod{}, which generates sequential actions considering an equitable workload among other agents. The effectiveness of \ourmethod{} is demonstrated through its superior performance in two representative min-max routing tasks: the min-max multiple traveling salesman problem (min-max mTSP) and the min-max multiple pick-up and delivery problem (min-max mPDP). Notably, our method achieves significant reductions of runtime, approximately 335 times, and cost values of about 53\% compared to a competitive heuristic (LKH3) in the case of 100 vehicles with 1,000 cities of mTSP. We provide reproducible source code: \url{https://anonymous.4open.science/r/equity-transformer}.

Min-max routing problems aim to minimize the maximum tour length among multiple agents 
by having agents conduct tasks in a cooperative manner.
% as they collaboratively visit all cities, i.e., the completion time. 
These problems include impactful real-world applications but are known as NP-hard. Existing methods are facing challenges, particularly in large-scale problems that require the coordination of numerous agents to cover thousands of cities. 
This paper proposes \ourmethod{} to solve large-scale min-max routing problems. \changed{First, we employ sequential planning approach to address min-max routing problems, allowing us to harness the powerful sequence generators (e.g., Transformer).}
% First, we model min-max routing problems into sequential planning, reducing the complexity and enabling the use of powerful Transformer architectures. 
Second, we propose key inductive biases that ensure equitable workload distribution among agents. The effectiveness of \ourmethod{} is demonstrated through its superior performance in two representative min-max routing tasks: the min-max multi-agent traveling salesman problem (min-max mTSP) and the min-max multi-agent pick-up and delivery problem (min-max mPDP).
% This paper proposes \ourmethod{} to solve large-scale min-max routing problems. \ourmethod{} ensures an equitable assignment of workload among various agents, effectively transforming a previously promising transformer-based combinatorial solver into a formidable solution for the min-max routing problem. The effectiveness of \ourmethod{} is demonstrated through its superior performance in two representative min-max routing tasks: the min-max multiple traveling salesman problem (min-max mTSP) and the min-max multiple pick-up and delivery problem (min-max mPDP). 
Notably, our method achieves significant reductions of runtime, approximately 335 times, and cost values of about 53\% compared to a competitive heuristic (LKH3) in the case of 100 vehicles with 1,000 cities of mTSP. We provide reproducible source code: \url{https://github.com/kaist-silab/equity-transformer}.
\end{abstract}

\section*{Introduction}

Routing problems are combinatorial optimization problems that are notoriously difficult to solve. The traveling salesman problem (TSP) and vehicle routing problems (VRPs) are representative problems where the objective is to determine the optimal or shortest tour route(s) for one or multiple agents, such as robots, vehicles, or drones. These problems are classified as NP-hard, posing significant challenges \citep{PAPADIMITRIOU1977237}. Various approaches have been proposed to solve routing problems, including mathematical programming techniques that aim to achieve provable optimality \citep{gurobi, concorde}, task-specific heuristic solvers \citep{lkh2017,ortools}, and deep learning-based methods that provide task-agnostic and fast heuristic solvers \citep{khalil,kool2018attention}. The deep learning-based methods have shown promising results even for large-scale problems that have more than 2,000 cities \citep{fu,qiu2022dimes,sun2023difusco,zhang2023neural, pmlr-v202-sun23c}.

% \cite{park2021schedulenet,kool2018attention,mis,khalil,bello2017neural,li2018combinatorial,duedon,Nazari,NLNS,chen2019learning,ma2019combinatorial,drl-2opt,kwon2020pomo,barrett2020exploratory,wu2020learning,xin2021multi,yoon2021deep,kwon2021matrix,kim2021learning,kim2022sym,ma2022efficient,qiu2022dimes}

% The traveling salesman problem (TSP) is a challenging combinatorial optimization problem that involves finding the shortest possible route for an agent (such as a robot, vehicle, or drone) to visit every city and return to the starting point (depot). TSP has several variants which have high-impact applications, pick-up and delivery (PDP) problem \cite{mosheiov1994travelling}, but it is proven to be NP-hard \cite{PAPADIMITRIOU1977237}. Several works have tried to tackle the NP-hard nature of TSP and its' variant; this includes mathematical programming, which focuses on provable optimality \cite{gurobi}, heuristic, which is a task-specialized fast approximated solver \cite{concorde,lkh2017,ortools}, and deep learning-based methods that aim task-agnostic fast approximate solver \cite{park2021schedulenet,kool2018attention,mis,khalil,bello2017neural,li2018combinatorial,duedon,Nazari,NLNS,chen2019learning,ma2019combinatorial,drl-2opt,kwon2020pomo,barrett2020exploratory,wu2020learning,xin2021multi,yoon2021deep,kwon2021matrix,kim2021learning,kim2022sym,ma2022efficient,qiu2022dimes}. These methods give promising performances even at large-scale problems which have cities of $N>10000$ \cite{concorde,fu2020generalize}. 

Min-max routing problems are distinct from standard (min-sum) routing problems in that they focus on minimizing the cost of the most expensive route among multiple agents \changed{(i.e., minimizing the total completion time)}, rather than minimizing the sum of the costs of routes. These problems are particularly relevant in time-critical applications such as disaster management \citep{cheikhrouhou2021comprehensive}, where minimizing completion time (or service time) is crucial. 
However, min-max routing problems are far more challenging than min-sum counterparts because algorithms for min-max routing require coordinated cooperation among multiple agents to ensure an equitable assignment of workload among them. Classical exact algorithms struggle to solve min-max routing problems due to their NP-hardness \citep{francca1995m}. Additionally, powerful heuristic approaches for min-sum problems are not well generalized to the min-max case \citep{bertazzi2015min}, particularly for large-scale problems \citep{nce}, owing to the inherent differences between min-max and min-sum problems.

% Recently, deep learning methods have been utilized to address min-max routing problems \citep{hu2020reinforcement,cao2021dan,park2021schedulenet} as an alternative to classical approaches. Notably, representative min-max routing techniques such as ScheduleNet \citep{park2021schedulenet} and the decentralized attention network (DAN) \citep{cao2021dan} aim to handle the min-max nature by treating it as a simultaneous decision-making process with decentralized modeling of multiple agents. These approaches capture the intuitive real-time execution of decision-making in multi-agent min-max routing; they modeled even-based execution directly, i.e., they directly model simultaneous execution depicted in Fig 1. Despite their intuitive decentralized modeling, the performances of these models are not competitive due to the difficulty of training decentralized cooperative policy using sparse and delayed reward \citep{park2021schedulenet}.

Recently, deep learning methods have been utilized to address min-max routing problems \citep{hu2020reinforcement,cao2021dan,park2021schedulenet} as an alternative to classical approaches. Notably, representative min-max routing techniques such as ScheduleNet \citep{park2021schedulenet} and the decentralized attention network (DAN) \citep{cao2021dan} aim to handle the min-max nature with event-based \textit{parallel planning}. They model a parallel decision-making process among multiple agents in a decentralized way. The parallel planning methods can be directly applied as a real-time dispatcher, which is advantageous in handling dynamic situations where states contain stochastic changes. 
% Nevertheless, parallel planning with decentralized agents has high modeling complexity, posing challenges when applied to large-scale \changed{routing} problems \citep{park2021schedulenet}.

% To tackle the limitation of parallel planning-based methods, we introduce \textit{sequential planning} method with a novel architecture, \ourmethod{}. We advocate for a fully centralized approach to effectively solve min-max routing problems, viewing them as instances of sequential planning rather than parallel planning. Specifically, we tackle the min-max routing problem by generating one long sequences, with each sub-sequence representing the tour for a specific agent. This perspective is driven by the modeling simplicity and the inherent flexibility of using powerful sequential architectures.

% To accomplish this, we employ a Transformer \citep{transformer,kool2018attention} model. On top of the Transformer, we introduce two essential inductive biases for min-max routing, which guide relational decision-making and ensure equitable workload assignment among the agents within the generated sequences: 

\begin{figure}
    \centering
    \includegraphics[width=\linewidth]{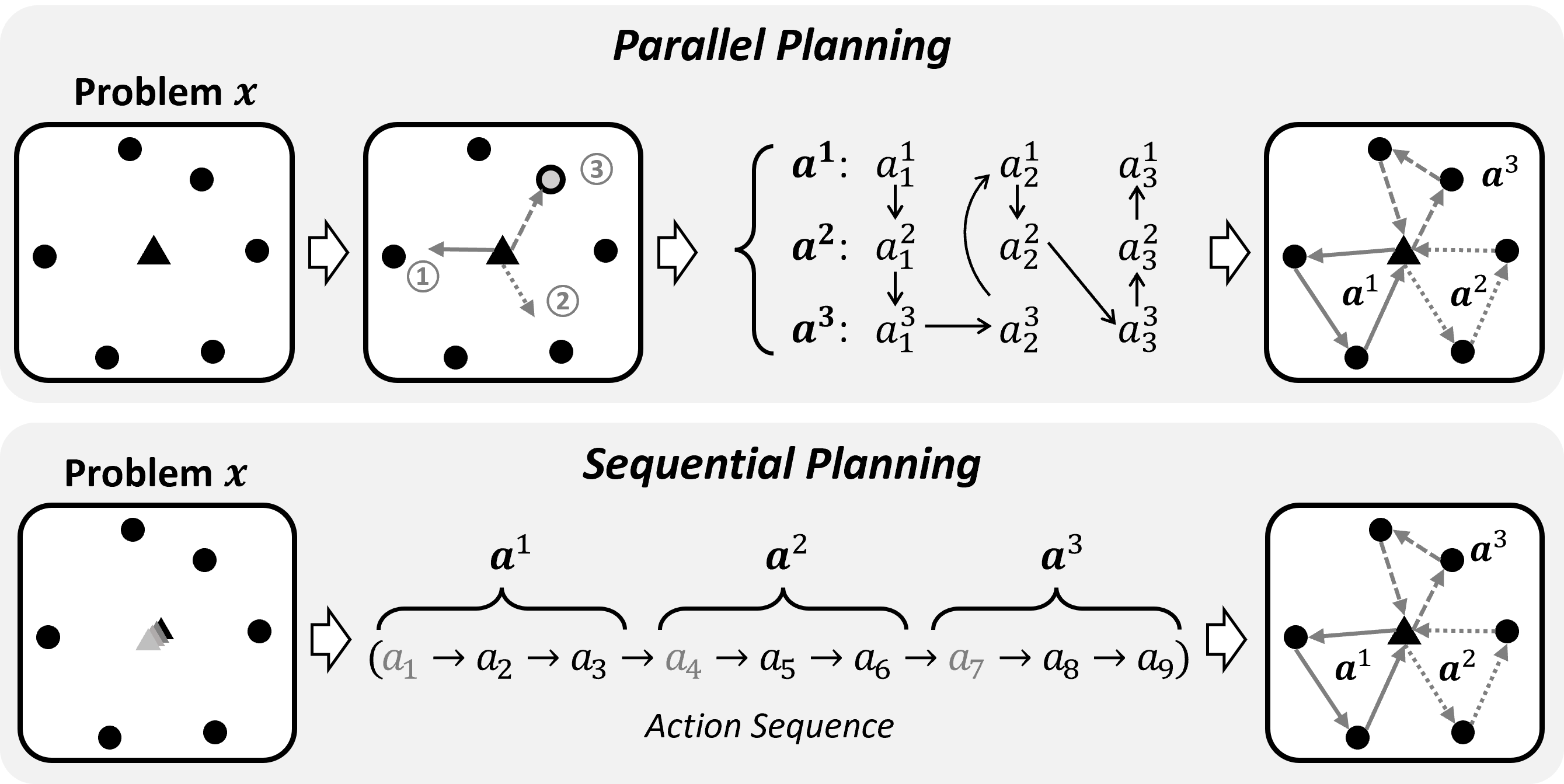}
    \caption{Illustration of parallel planning and sequential planning on min-max VRPs. In sequential planning, 
    when an action selects a depot index (gray colored $a_t$), the planning for currently active agent is terminated to start planning for the new agent that corresponds to the selected depot.
    }
    \label{fig:planning}
\end{figure}

% The \changed{min-max routing problems} can be modeled in \textit{sequential planning} by introducing artificial orders among agents. \Cref{fig:planning} shows the difference between parallel planning and sequential planning on VRPs. 
\changed{
% However, parallel planning encounters a notable challenge in the form of high modeling complexity, necessitating decisions to be made within the joint space of agent and city selections. 
However, parallel planning encounters challenges in modeling decentralized decision-making, which necessitates searching for the joint space of agents' actions. These challenges become particularly pronounced when attempting to apply parallel planning to large-scale routing problems \citep{park2021schedulenet}.
On the contrary, sequential planning presents an alternative approach involving a hierarchical decomposition of action choices among agents. This results in a substantial reduction in modeling complexity when compared to parallel planning. However, an important drawback of sequential planning lies in its diminished relational context between agents due to its sequential representation, which might lead to imbalanced tours among agents, i.e., increased tour costs. \Cref{fig:planning} illustrates the difference between parallel planning and sequential planning.}

% Conversely, sequential planning presents an alternative approach involving a hierarchical decomposition of action choices among agents. This results in a substantial reduction in modeling complexity when compared to parallel planning. However, an important drawback of sequential planning lies in its diminished relational context between agents due to its sequential representation, which has the potential to affect min-max routing performance adversely.}

\changed{To achieve equitable assignment and keep leveraging reduced complexity via sequential planning, we propose a novel sequential planning architecture, \emph{\ourmethod{}}.
% We advocate for a fully centralized approach to effectively solve min-max routing problems, viewing them as instances of sequential planning rather than parallel planning. 
Specifically, we tackle the min-max routing problems by generating one long sequence via \ourmethod{}, where each sub-sequence represents a specific agent's tour.
To contextualize relational decision-making and ensure equitable workload assignment among the agents,
\ourmethod{} introduces two essential inductive biases as follows:
% \ourmethod{} introduces two essential inductive biases to compare with the original transformer \cite{transformer} to contextualize relational decision-making and ensure equitable workload assignment among the agents when the action sequences are generated. The proposed inductive biases are as follows: 
}

% Simplified modeling enables the employment of powerful sequential architectures, such as Transformer \citep{transformer,kool2018attention}

\begin{itemize}
    \item \textbf{Multi-agent positional encoding for order bias.} We introduce virtual orders on agents to model a parallel decision-making process as a sequence. 
    The homogeneous agents are modeled with the precedence (i.e., order bias among agents), we add positional encoding and inject it into the encoder.
    % To model the precedence of agents (i.e., order bias among agents), we add positional encoding for ordered agents and inject it into the encoder.
    \item \textbf{Context encoder for equity.} To promote equitable tours for multiple agents, we incorporate an equity context into the sequence generator. Equity context considers the temporal tour length, the target tour length, and the desired number of cities to be visited, which are essential factors for enhancing the fairness of the generated tours.
\end{itemize}

Our method performs remarkably well at the min-max routing problems, outperforming both existing classical heuristic and learning-based methods. As a highlight, \ourmethod{} achieves 334$\times$ speed improvement and 53\% reduction of solution cost compared to the representative classical heuristic solver (LKH3) when solving the multi-agent TSP with 1,000 cities. Also, our method achieves 1,217$\times$ faster speed and 9\% reduction of solution cost than the representative learning-based method (ScheduleNet) with parallel planning.

\section*{Problem Formulation}

In our work, we focus on tackling min-max routing problems, which involve a scenario where a group of $M$ agents needs to visit $N$ cities with the objective of minimizing the maximum tour length among the individual tours of the agents. In this section, we present the formulation of min-max routing through the lens of sequential planning.
% (see \cref{fig:planning} for a detailed process).

\subsubsection{Problem.} 
A routing problem is defined the set of city locations and a depot, represented as $\bm{x}=\{x_i\}_{i=1}^{N}$ where $x_i \in \mathbb{R}^2$ is the Cartesian coordinates, and $x_1$ denotes the depot. Since the $M$ agents find tours that start from the depot and return to the depot, we add $M$ dummy depot (each dummy depot assigned to each agent), i.e., $x_{N+1}=\cdots=x_{N+M}=x_1$. Thus, the sequential planning routing is defined as $\bm{x}=\{x_i\}_{i=1}^{N+M}$.
Remarks that we can expand the definition of $\bm{x}$ so that it can include additional features required for other problems like the capacitated vehicle routing. For simplicity, we represent locations only.

\subsubsection{Action.} 
The sequential planning is represented as an action sequence $\bm{a}=(a_1, \ldots, a_{N+M})$. This action sequence is formed by selecting an index from the set of unvisited nodes at step $t$, i.e., $a_t \in \{1, \ldots, N+M\} \setminus \{a_1, \ldots, a_{t-1}\}$. 
The resulting action sequence is partitioned into $M$ subsequences, i.e., agent tours $(\bm{a}^1, \ldots, \bm{a}^M)$, by splitting $\bm{a}$ with depot choosing actions.  
Thus, each $\bm{a}^m = (a^m_{1}, \ldots, a^m_{L_m})$ starts with a dummy depot index, i.e., $a^m_{1} \in \{N+1, \ldots, N+M\}$, followed by subsequent city indices. 
Please refer to \Cref{fig:planning}.

\subsubsection{State.} The state $s_t$ is defined as the union of the precollected actions $a_1, \ldots, a_{t-1}$ and the problem $\bm{x}$, i.e., $s_1 = \{\bm{x}\}$, and $s_t = \{a_1, \ldots, a_{t-1};\bm{x}\}$ for $t>1$.

% \subsubsection{Solution.} The solution is terminal state: $\bm{s}_{N+M}$. 

% $\bm{a} = \{a_i\}_{i=1}^{N+M}$ is a permutation index sequence of the problem: i.e., $a_i \in \{1, ..., N+M\}$ and $a_i \neq a_j$ if $i \neq j$. Note that $a \in \{1, ..., N\}$ stands for the cities, whereas $a \in \{N+1, ..., N+M\}$ stands for the dummy depot (i.e., the virtual return point) for each agent. The whole solution $\bm{a}$ can be partitioned into $M$ sub-sequence, each of which represents the agent's tour, i.e., $\bm{a} = (a_1, \ldots, a_{N+M}) = (\bm{a}^1, \ldots, \bm{a}^M)$. 

\subsubsection{Cost.} The cost is the maximum tour length among all agents' tours of $(\bm{a}^1, \ldots, \bm{a}^M)$ of given action sequence $\bm{a}$, i.e.,
\begin{align*}
    & \mathcal{L}_{\text{cost}}(\bm{a};\bm{x}) := \max \left\{ \mathcal{L}(\bm{a}^1;\bm{x}),...,\mathcal{L}(\bm{a}^M;\bm{x})\right\}, \mbox{where}\\
     & \mathcal{L}(\bm{a}^m;\bm{x}) := \sum_{t = 2}^{L_m}  ||x_{a^m_t}-x_{a^m_{t-1}}||_{2} + ||x_{a^m_1}-x_{a^m_{L_m}}||_2.
\end{align*} 

% \textbf{Agent Tour.} The agent tour $\bm{k} = (k_1,...,k_L)$ is sub-sequence of solution $\bm{e}$ which has characteristic of $k_1 \in [N+1,M+N]$ (agent) and $k_2,...,k_{L} \in [N]$ (cities visited by the agent). The solution $\bm{e} =(\underbrace{k^1_1,...,k^1_{L_1}}_{\text{agent tour: } \bm{k}^1},...,\underbrace{k^M_1,...,k^M_{L_M}}_{\text{agent tour: }\bm{k}^M})$ can be decomposed as $M$ agent tour of $\bm{k}^1,...,\bm{k}^M \in \mathcal{K}$ where $\mathcal{K}$ be a set of all agent tours. 

% \textbf{Objective.} The objective is to minimize the maximum tour length among agents' tour $\bm{k} \in \mathcal{K}$:
% \begin{align*}
%     \mathcal{L}(\bm{k};\bm{x}) &:= \sum_{i = 1}^{L} ||x_{k_i}-x_{k_{i-1}}||_{2} + ||x_{k_1}-x_{k_{L}}||_2 \\
%     \mathcal{L}_{\text{obj}}(\bm{e};\bm{x}) &:= \max \left(\mathcal{L}(\bm{k}^1;\bm{x}),...,\mathcal{L}(\bm{k}^M;\bm{x})\right).
% \end{align*} 

\subsubsection{Policy.} 
The policy $\pi_{\theta}(\bm{a}|\bm{x})$ is a composition of segment policy $\pi_{\theta}(a_t|\bm{s}_t)$, generating action sequences for given problem condition $\bm{x}$ according to the following expression.
% The policy $\pi_{\theta}(\bm{a}|\bm{x})$ is composition of segment policy $\pi_{\theta}(a_t|\bm{s}_t)$ for generating action sequences $\bm{a} = (a_1, \ldots, a_{N+M})$ of given problem condition $\bm{x}$ as follows:
\begin{equation*}
    \pi_{\theta}(\bm{a}|\bm{x}) = \prod_{t=1}^{N+M}\pi_{\theta}(a_t|\bm{s}_t).
\end{equation*}
The $\theta$ is the deep neural network parameter of the policy $\pi$. 
\changed{The optimal parameter $\theta^*$ can be determined by solving the following optimization problem:}
\begin{equation*}
    \theta^* = \argmin_{\theta} \mathbb{E}_{P(\bm{x})}\mathbb{E}_{\pi_{\theta}(\bm{a}|\bm{x})} \mathcal L_{\text{cost}}(\bm{a};\bm{x}),
\end{equation*}
where $P(\bm{x})$ is the distribution of problem $\bm{x}$.

\section*{Methodology}

% This section presents the architecture of \ourmethod{} $\pi_{\theta}(\bm{a}|\bm{x})$ which generates action sequence $\bm{a} = \{a_1, \ldots, a_{N+M}\}$, for given problem $\bm{x}$. 
This section presents the architecture of \ourmethod{} $\pi_{\theta}(\bm{a}|\bm{x})$, which generates an action sequence $\bm{a} = (a_1, \ldots, a_{N+M})$ for a given problem $\bm{x}$.
Our high-level idea is to build a transformer model with \textit{multi-agent positional encoding} and \textit{equity context}. 

Our architecture has the following forward propagation:

\begin{enumerate}
    \item Multi-agent positional encoding for initial node embedding given problem $\bm{x}$.
    \item Employ the encoder of Transformer \citep{transformer} to the initial node embedding to obtain $\bm{H} =[h_1,\ldots,h_{N+M}] \in \mathbb{R}^{D \times (N+M)}$, where $D$ is the embedding dimension.
    \item Iterative decoding $t=1,\ldots,N+M$ using
    \begin{enumerate}
        \item Equity context encoding for $\bm{c}_t \in \mathbb{R}^{D}$.
        \item Decoding to produce $\bm{a}_t \sim \pi_{\theta}(\bm{a}_t|\bm{H}, \bm{c}_t)$ by using $c_t$ as attention query of decoder.    
    \end{enumerate} 
\end{enumerate}

% The encoding procedure and iterative decoding with contextual queries have been extensively covered in the existing Transformer literature and AM framework by \citet{kool2018attention}. 
% In this paper, our primary focus lies in introducing our new elements: the multi-agent positional encoding and the equity context encoding.
The encoding and the iterative decoding procedure process involving contextual queries have been comprehensively addressed in the previous literature \citep{kool2018attention, kwon2020pomo, li2021heterogeneous, kim2022sym}. 
In this paper, we focus on introducing new elements designed for min-max routing problems, which are multi-agent positional encoding and equity context encoding.

% The encoding procedure, as well as the iterative decoding process involving contextual queries, have been comprehensively addressed in the extant Transformer literature. In this paper, our primary focus lies in introducing our new elements, specifically the multi-agent positional encoding and the equity context encoding for min-max routing problems.

\subsection*{Multi-agent Positional Encoding} \label{sec:mpe}

We begin with partitioning the problem $\bm{x}$ into two distinct components: the cities, denoted as $\bm{x}_{\text{city}}$ with elements $\{x_i\}_{i=1}^{N}$, and the agents, represented by $\bm{x}_{\text{agent}}$ with elements $\{x_i\}_{i=N+1}^{N+M}$. 
In order to facilitate sequential relationships between the agents, we employ positional encoding $f_{\text{PE}}$ to $\bm{x}_{\text{agent}}$. This positional encoding incorporates sine and cosine functions with differing frequencies, following the work by \citet{transformer}.

% We begin by decomposing the problem represented by $\bm{x}$ into two distinct parts: the cities, denoted as $\bm{x}_{\text{city}}$ with elements $\{x_i\}_{i=1}^{N}$, and the agents, represented by $\bm{x}_{\text{agent}}$ with elements $\{x_i\}_{i=N+1}^{N+M}$. To enable effective positional awareness within the $\bm{x}_{\text{agent}}$, we employ positional encoding $f_{\text{PE}}$, as described in \citet{transformer}, which incorporates sine and cosine functions with differing frequencies.

% Next, we concatenate the linearly projected vectors of $\bm{x}_{\text{city}}$ and $f_{\text{PE}}(\bm{x}_{\text{agent}})$ to form the initial node embedding. The embedding is subsequently fed into the encoder, a structural component akin to the attention model (AM) by \citet{kool2018attention}. This encoder includes multi-head attention, batch normalization, and a feed-forward network, as illustrated in \Cref{fig:full_architecture}. This architecture ensures that our model possesses the necessary positional and structural awareness for efficient problem-solving.

\begin{figure}[t!]
\centering

\includegraphics[width=0.45\textwidth]{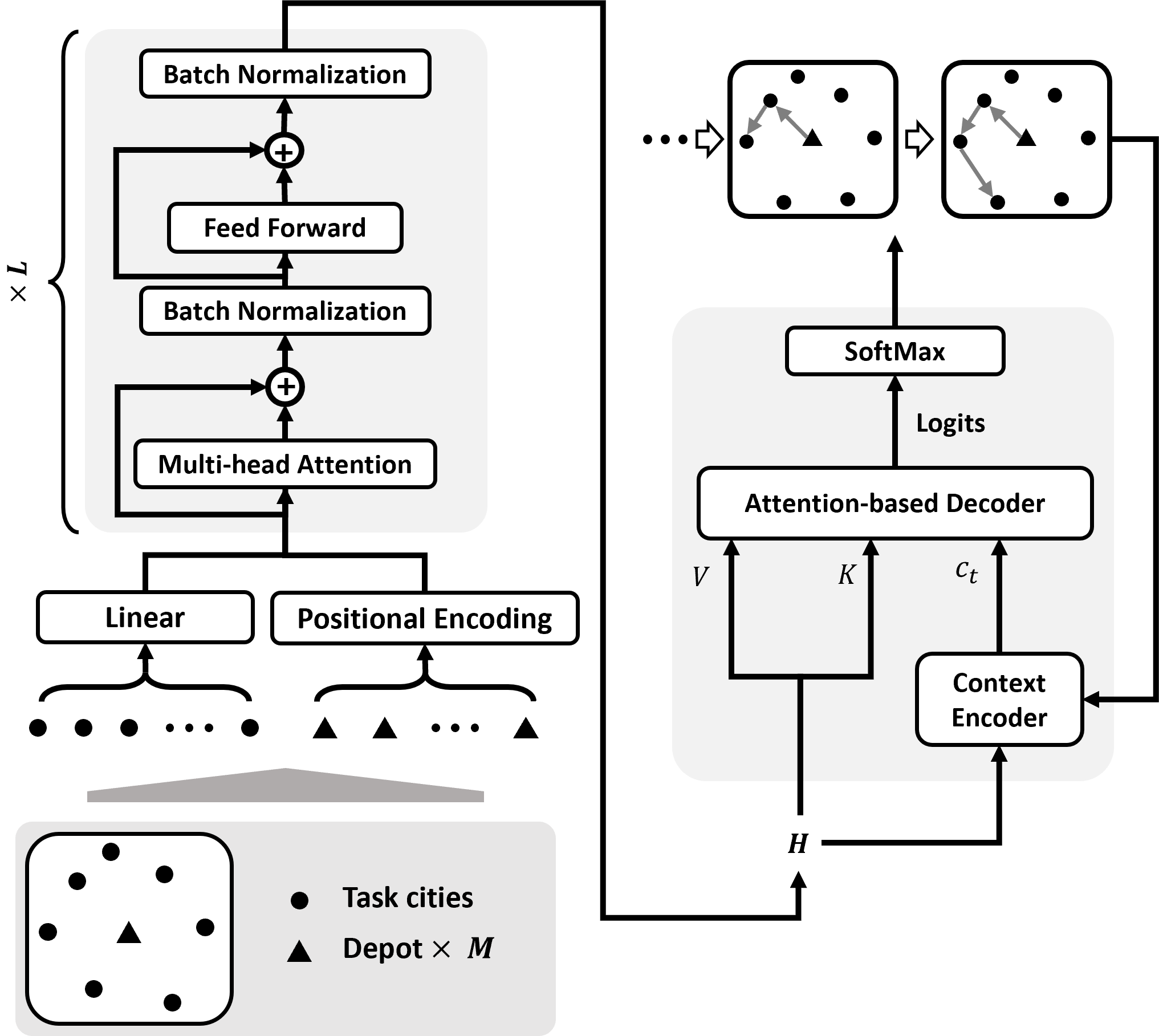}

\caption{
Illustration of \ourmethod{}. The $L$ stands for the number of sequential layers, where we set $L=3$.}

\label{fig:full_architecture}
\end{figure}

Next, we concatenate the linearly projected vectors of $\bm{x}_{\text{city}}$ and $f_{\text{PE}}(\bm{x}_{\text{agent}})$ to form the initial node embedding. The embedding is subsequently fed into the encoder, a structural component akin to the encoder of AM as illustrated in \Cref{fig:full_architecture}.
% A significant departure from the original AM lies in their disregard for positional encoding. Their approach assumes that $\bm{x}$ represents a set rather than a sequence, rendering positional encoding less relevant due to the absence of inherent sequential bias.
It is worth to note that the original AM architecture itself is not appropriate for addressing multi-agent problems, due to its incapability to consider the multi-agent nature.
In contrast, our approach incorporates positional encoding to $\bm{x}_{\text{agent}}$ with the specific intention of introducing a virtual order bias. Consequently, we can sequentially generate the tour sequence of an agent while considering both preceding and succeeding agents in the assigned order.

\begin{figure}[t!]
\centering

% \usepackage{caption}
% \captionsetup[figure]{font=small}

\includegraphics[width=0.47\textwidth]{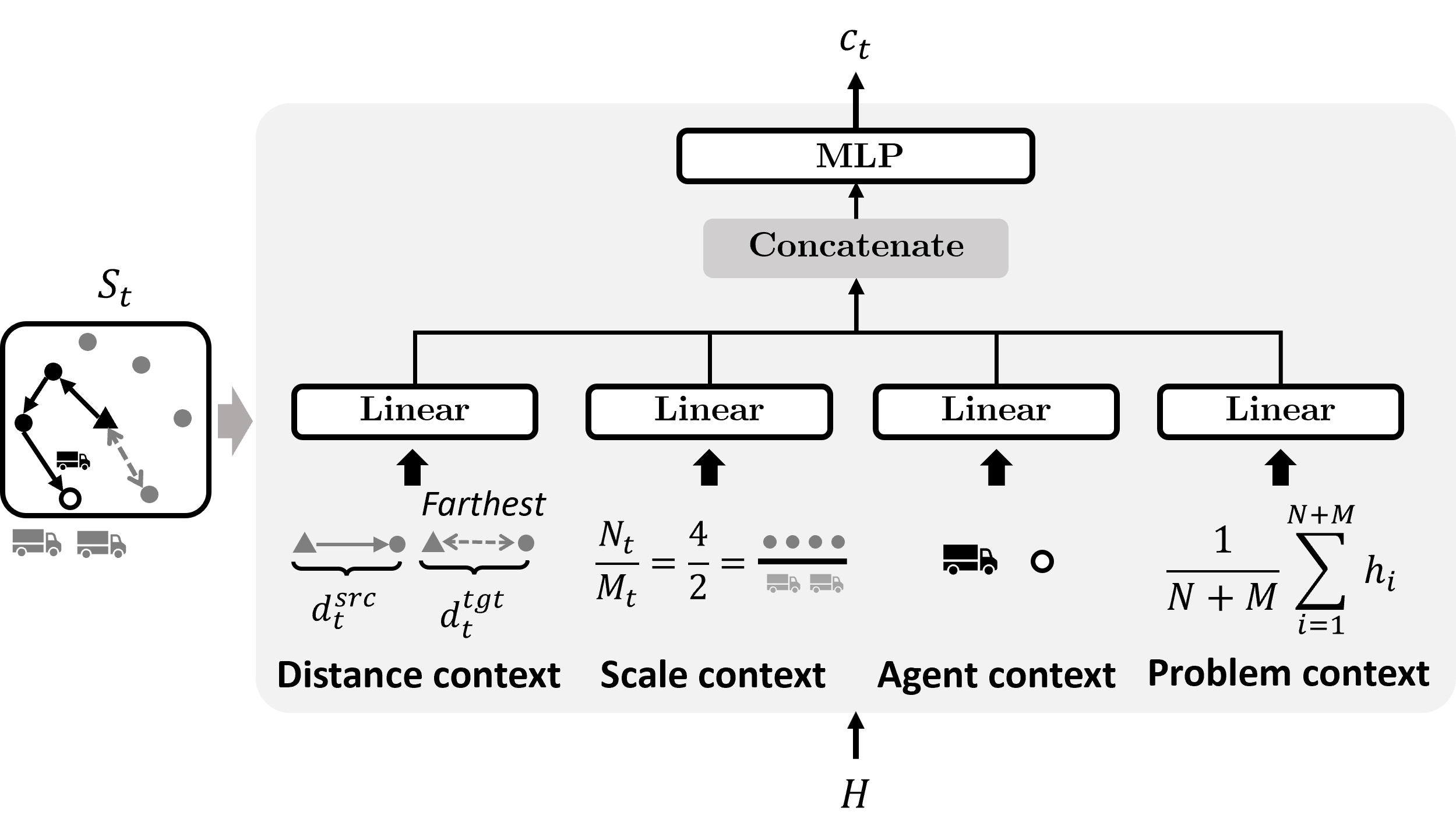}
\vspace{-5pt}
\caption{Illustration of equity context encoding}
\label{fig:context_enc}
\end{figure}

\subsection*{Equity Context Encoding}\label{sec:con-en}
For every decoding step $t$, we utilize four distinct contexts as ingredients of the equity context, $\bm{c}_t$. Each of the ingredient contains useful information for equitable decoding, described as follows:

%Our context encoder $f_{\text{CE}}:\mathbb{R}^{4D} \rightarrow \mathbb{R}^{D}$, which is a multi-layer perception (MLP), produces equity context $\bm{c}_t \in \mathbb{R}^{D}$ to be used for decoding (i.e., generating) multiple balanced tours. We introduce four contexts at the current step $t$ as follows: 

% (1) problem context $\bm{h}^{\text{problem}} \in \mathbb{R}^{D}$, (2) agent context $\bm{h}^{\text{agent}}_t \in \mathbb{R}^{D}$, (3) scale context $\bm{h}^{\text{scale}}_t \in \mathbb{R}^{D}$, and (4) distance context $\bm{h}^{\text{dist}}_t \in \mathbb{R}^{D}$. 

\begin{itemize}
    \item \textbf{Problem context $\bm{h}^{\text{problem}} \in \mathbb{R}^{D}$:} The problem context collects the average of representations \\
    $\bm{h}^{\text{problem}} =  \frac{1}{M+N}\sum_{i=1}^{M+N}\bm{h}_{i}$. This aligns with the context embedding of AM, which is primarily intended to capture the global context of problem $\bm{x}$ by averaging each city and agent representation.

    \item \textbf{Agent context $\bm{h}^{\text{agent}}_t \in \mathbb{R}^{D}$:} We set agent context using representations of the returning depot and the last visited city of the current active agent $m_t$.
    % We set the agents context with the representation of the current active agent $m_t$, i.e., 
    % $\bm{h}^{\text{agent}}_t = \bm{h}_{N+m_t}$.
    % $\bm{h}_{\text{acting}} = \bm{h}_{i_t}$, where $i_t$ denotes the index of the active agent at step $t$, i.e., $i_t \in \{N+1,...,N+M\}$. 
    % Additionally, we gather the representation of the city which the active agent is currently visiting: $\bm{h}_t$, where $t \in \{1,...,N\}$.
    Precisely, $\bm{h}^{\text{agent}}_t = g_{\text{agent}}(\bm{h}_{N+m_t} \oplus \bm{h}_{a_{t-1}})$, where $g_{\text{agent}}:\mathbb{R}^{2D} \rightarrow \mathbb{R}^{D}$ represents a linear projection. This context highlights the currently active agents.

    \item \textbf{Scale context $\bm{h}^{\text{scale}}_t \in \mathbb{R}^{D}$:} We incorporate the ratio between $N_t$ and $M_t$ as a scaling context, where $N_t$ represents the number of remaining cities and $M_t$ denotes the current number of un-used agents at the depot, i.e., $N_t/M_t \in \mathbb{R}$. We then generate $\bm{h}^{\text{scale}}_t = g_{\text{scale}}\left(N_t/M_t\right)$, where $g_{\text{scale}}:\mathbb{R} \rightarrow \mathbb{R}^{D}$ represents a linear projection. This context offers valuable insights into the approximate number of cities an agent should visit to achieve equity. Consequently, the scale ratio can provide the effective information to decide whether to continue visiting additional cities or return to the depot.

    \item \textbf{Distance context $\bm{h}^{\text{dist}}_t \in \mathbb{R}^{D}$:} \recheck{We make use of dynamic changes in the agent's tour length and the distance of remaining cities from the depot at the current step $t$}. Firstly, we employ $d_t^{\text{source}}$, which represents the current tour length of the active agent. Secondly, we utilize $d_t^{\text{target}}$, which denotes the maximum distance between the depot and the remaining unvisited cities. Subsequently, we form $\bm{h}^{\text{dist}}_t = g_{\text{dist}}(d_t^{\text{source}} \oplus d_t^{\text{target}})$, where $g_{\text{dist}}:\mathbb{R}^2 \rightarrow \mathbb{R}^{D}$ is a linear projection. This information holds significant importance in terms of the equity of tour length among agents in the min-max routing problem. The context prompts the decoders to consider the agent's current tour length and remaining tasks, aiding in the decision-making process of whether to stop visiting (i.e., return to the depot) or continue the tour while considering the min-max tour length.
\end{itemize}

%These four contexts form a valuable source of task-equity information for multiple agents \changed{ so that \ourmethod{} sequentially produces the balanced tours} while taking into account the remaining tasks of other agents. We aggregate every above context and produce $\bm{c}_t$ via $f_{\text{CE}}$, i.e., 

%Then, we use $\bm{c}_t$ as a contextual query for constructing policy $\pi_{\theta}(\bm{a}_t|\bm{H}, \bm{c}_t)$ by using multi-head attention process as shown in \Cref{fig:context_enc}.

The context encoder $f_{\text{CE}}:\mathbb{R}^{4D} \rightarrow \mathbb{R}^{D}$, which is a multi-layer perception (MLP), produces equity context $\bm{c}_t$ using these four contexts, i.e.,
\begin{equation*}
    \bm{c}_t = f_{\text{CE}}(\bm{h}^{\text{problem}} \oplus \bm{h}^{\text{agent}}_t \oplus \bm{h}^{\text{scale}}_t \oplus \bm{h}^{\text{dist}}_t).
\end{equation*}
We adopt an approach similar to that of \citet{kool2018attention}, where use $\bm{c}_t$ as a contextual query for the attention-based decoder $\pi_{\theta}(\bm{a}_t|\bm{H}, \bm{c}_t)$, as shown in \Cref{fig:context_enc}. This utilization of task-equity information from the equity context enables the decoder to sequentially generate balanced tours.

\subsection*{Training Scheme}

The \ourmethod{} is trained with REINFORCE \citep{williams1992simple} with the shared baseline scheme similar to \citet{kwon2020pomo} and \citet{kim2022sym}. The shared baseline with symmetric samples makes symmetric exploration for the combinatorial solution space. The training loss with the symmetric shared baselines is as follows:
\begin{equation*} \label{eq1}
\begin{split}
    \mathcal{L}_{\text{train}}(\bm{\theta}) &=\mathbb{E}_{P(\bm{x})}\mathbb{E}_{\pi_{\theta}(\bm{a}|\bm{x})} \mathcal{L}_{\text{cost}}(\bm{a};\bm{x}), \\
    \nabla \mathcal{L}_{\text{train}}(\bm{\theta})
    & \approx \sum_{i=1}^B \sum_{j=1}^L \left(\mathcal{L}_{\text{cost}}\left(\bm{a}^{(i,j)};\bm{x}^{(i)}\right)- b_{\text{shared}}\right),
\end{split}
\end{equation*}
where $b_{\text{shared}}:=1/L\sum_{j=1}^{L}\mathcal{L}_{\text{cost}}(\bm{a}^{(i,j)};\bm{x}^{(i)})$. 
% \begin{align*}
%     &\mathcal{L}_{\text{train}}(\bm{\theta}) =\mathbb{E}_{P(\bm{x})}\mathbb{E}_{\pi_{\theta}(\bm{a}|\bm{x})} \mathcal{L}_{\text{cost}}(\bm{a};\bm{x}), \\
%     &\nabla \mathcal{L}_{\text{train}}(\bm{\theta}) \\
%     & \quad \approx \sum_{i=1}^B \sum_{j=1}^L \left(\mathcal{L}_{\text{cost}}\left(\bm{a}^{(i,j)};\bm{x}^{(i)}\right)- b \left( \bm{a}^{(i,1)},...,\bm{a}^{(i,L)} \right)\right).
% \end{align*}
Each $\bm{a}^{(i,l)}$ is sampled sequence from training solver given $L$ symmetric $\bm{x}^{(i)}$: $\pi_{\theta}(\bm{a}|\mathcal{T}_1(\bm{x}^{(i)})),...,\pi_{\theta}(\bm{a}|\mathcal{T}_L(\bm{x}^{(i)}))$, where $\mathcal{T}_1,...,\mathcal{T}_L$ are symmetric transformation of problem instance $\bm{x}^{(i)}$. 
% The $b_{\text{shared}}$ is shared baseline which is average value of $L_{\text{cost}}$ among $\bm{a}^{(i,1)},...,\bm{a}^{(i,L)}$: $\frac{1}{L}\sum_{j=1}^{L}\mathcal{L}_{\text{cost}}(\bm{a}^{(i,j)};\bm{x}^{(i)})$. 
See \citet{kim2022sym} for a detailed training scheme.

\section*{Experiments}

\begin{table*}[t!]
\begin{center}
\vspace{0.01in}
\resizebox{0.86\linewidth}{!}{
\begin{tabular}{ccccccccc}
\toprule
&
& \multicolumn{4}{c}{Classic-based}
& \multicolumn{3}{c}{Learning-based}
%& \multicolumn{1}{c}{Ours}
\\
\cmidrule(lr{0.2em}){3-6}
\cmidrule(lr{0.2em}){7-9} 
% \cmidrule(lr{0.2em}){9-9}
\multicolumn{1}{c}{$N$}
& \multicolumn{1}{c}{$M$} 
& \multicolumn{1}{c}{LKH3 (60)}
& \multicolumn{1}{c}{LKH3 (600)}
& \multicolumn{1}{c}{OR-Tools (60)}
& \multicolumn{1}{c}{OR-Tools (600)}
& \multicolumn{1}{c}{SN }
& \multicolumn{1}{c}{NCE }
% & \multicolumn{1}{c}{MF (N.C.)}
& \multicolumn{1}{c}{ET (\textit{ours})}
\\
\midrule
\multirow{4}{*}{\shortstack[l]{200}}
&10
&2.52 \small{(60)}
&2.08 \small{(600)}
&4.97 \small{(60)}
&2.22 \small{(600)}
&2.35 \small{(9.70)}
&2.07 \small{(5.07)}
% &2.15 \small{(0.03)}
&\textbf{2.05} \small{(0.36)}

\\
\addlinespace
&15
&2.39 \small{(60)}
&2.03 \small{(600)}
&4.82 \small{(60)}
&2.15 \small{(600)}
&2.13 \small{(10.52)}
&1.97 \small{(5.07)}
% &1.99 \small{(0.03)}
&\textbf{1.97} \small{(0.37)}

\\
\addlinespace
&20 
&2.29 \small{(60)}
&2.02 \small{(600)}
&3.74 \small{(60)}
&2.04 \small{(600)}
&2.07 \small{(11.40)}
&1.96 \small{(5.07)}
% &1.98 \small{(0.03)}
&\textbf{1.96} \small{(0.37)}

\\
\midrule
\multirow{4}{*}{\shortstack[l]{500}}
&30
&3.31 \small{(60)}
&2.70 \small{(600)}
&7.90 \small{(60)}
&6.44 \small{(600)}
&2.16 \small{(171)}
&2.07 \small{(5.20)}
% &2.40 \small{(0.05)}
&\textbf{2.02} \small{(0.87)}

\\
\addlinespace
&40
&3.10 \small{(60)}
&2.55 \small{(600)}
&7.46 \small{(60)}
&6.69 \small{(600)}
&2.12 \small{(276)}
&2.01 \small{(5.38)}
% &2.22 \small{(0.05)}
&\textbf{2.01} \small{(0.90)}

\\
\addlinespace
&50 
&2.93 \small{(60)}
&2.48 \small{(600)}
&8.50 \small{(60)}
&7.26 \small{(600)}
&2.09 \small{(217)}
&2.01 \small{(5.05)}
% &2.14 \small{(0.05)}
&\textbf{2.01} \small{(0.92)}

\\
\midrule
\multirow{4}{*}{\shortstack[l]{1000}}
&50
&4.45 \small{(60)}
&3.77 \small{(600)}
&11.65 \small{(60)}
&9.89 \small{(600)}
&2.26 \small{(2094)}
&2.13 \small{(15.05)}
% &3.03 \small{(0.13)}
&\textbf{2.06} \small{(1.72)}

\\
\addlinespace
&75
&3.71 \small{(60)}
&3.26 \small{(600)}
&13.16 \small{(60)}
&11.50 \small{(600)}
&2.17 \small{(1678)}
&2.07 \small{(15.05)}
% &2.39 \small{(0.14)}
&\textbf{2.05} \small{(1.80)}

\\
\addlinespace
&100
&3.23 \small{(60)}
&2.92 \small{(600)}
&10.79 \small{(60)}
&8.93 \small{(600)}
&2.16 \small{(1588)}
&2.05 \small{(15.01)}
% &2.30 \small{(0.14)}
&\textbf{2.05} \small{(1.79)}

\\
\midrule
\multirow{4}{*}{\shortstack[l]{2000}}
&100
&6.60 \small{(60)}
&4.61 \small{(600)}
&20.99 \small{(60)}
&18.85 \small{(600)}
&\textit{OB}
&2.85 \small{(43.96)}
% &2.40 \small{(0.05)}
&\textbf{2.09} \small{(3.49)}

\\
\addlinespace
&150
&5.08 \small{(60)}
&4.02 \small{(600)}
&14.00 \small{(60)}
&13.17 \small{(600)}
&\textit{OB}
&2.83 \small{(44.77)}
% &2.22 \small{(0.05)}
&\textbf{2.08} \small{(3.41)}

\\
\addlinespace
&200
&4.13 \small{(60)}
&3.36 \small{(600)}
&11.00 \small{(60)}
&10.41 \small{(600)}
&\textit{OB}
&2.08 \small{(30.30)}
% &2.14 \small{(0.05)}
&\textbf{2.08} \small{(3.60)}

\\
\midrule
\multirow{4}{*}{\shortstack[l]{5000}}
&300
&12.30 \small{(60)}
&7.87 \small{(600)}
&17.00 \small{(60)}
&17.00 \small{(60)}
&\textit{OB}
&2.97 \small{(290)}
% &3.03 \small{(0.13)}
&\textbf{2.40} \small{(8.78)}

\\
\addlinespace
&400
&8.85 \small{(60)}
&6.15 \small{(600)}
&13.00 \small{(60)}
&13.00 \small{(600)}
&\textit{OB}
&2.92 \small{(204)}
% &2.39 \small{(0.14)}
&\textbf{2.21} \small{(8.61)}

\\
\addlinespace
&500
&7.14 \small{(60)}
&5.37 \small{(600)}
&11.00 \small{(60)}
&11.00 \small{(600)}
&\textit{OB}
&2.89 \small{(198)}
% &2.30 \small{(0.14)}
&\textbf{2.19} \small{(9.02)}

\\
\bottomrule
\end{tabular}}%
\end{center}
\caption{
{Results on min-max mTSP.  Every performance is average performance among 100 instances. The bold symbol indicates the best performance. Average running times (in seconds) are provided in brackets. 
}
}
\label{tab:mtsp}
\vspace{-5pt}
\end{table*}

\begin{figure*}[t!]
\centering
\begin{subfigure}{.40\textwidth}
\centering
\includegraphics[width=0.8\textwidth]{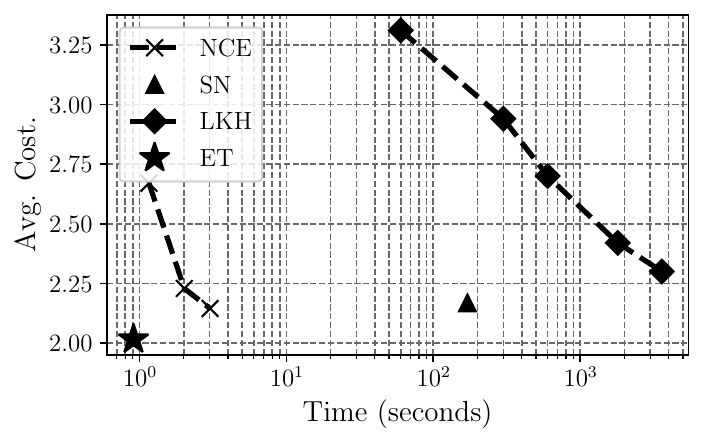}

\caption{mTSP with $(500, 30)$}
\end{subfigure}
\hspace{0.2in}
\begin{subfigure}{.40\textwidth}
\centering
%\vspace{0.0in}
\includegraphics[width=0.8\textwidth]{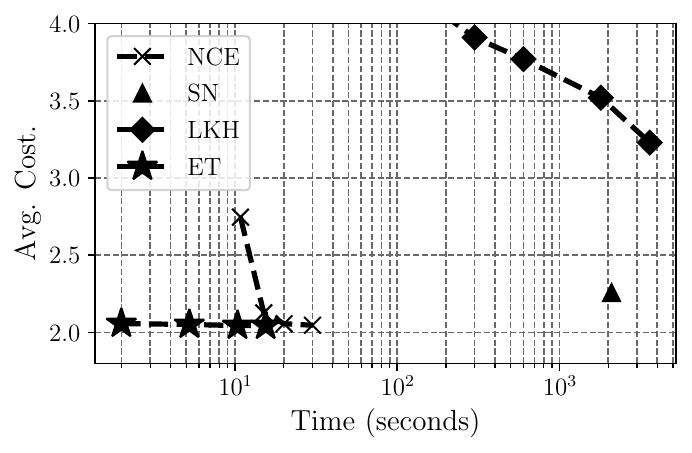}

\caption{mTSP with $(1000, 50)$}
\end{subfigure}
\caption{
Time-performance trade-off graph for mTSP. The left and bottom indicate the Pareto frontier. 
}
\label{fig:tradeoff_mtsp}
\end{figure*}

In this section, we present the experimental results of the \ourmethod{} model on two min-max routing problems: the multi-agent traveling salesman problem (mTSP) and the multi-agent pick-up and delivery problem (mPDP).

\subsubsection{Training Setting.} we use uniform distribution for the problem distribution $P(\bm{x})$, following \citet{kool2018attention}. For the training hyperparameters we set exactly the same hyperparameters for every task and experiment; see Appendix B. We train \ourmethod{} on $N=50$, and finetune it to target distribution of $N=200, 500$. 
The training time for the min-max mTSP is approximately one day, while for the min-max mPDP, it takes around four days.

\subsubsection{Experiments Metric.} It is important to carefully measure the performance comparison between methods, as often there is a trade-off between run time and solution quality. To this end, we present time-performance multi-objective graphs to compare tradeoffs between performance and computation time. In the result tables, we present the average cost achieved within a specific time limit, recognizing that every method has the potential to reach optimality given an unlimited amount of time.

\subsubsection{Target Problem Instances.} To evaluate the performance of our methods, we report results on randomly generated synthetic instances of min-max mTSP and min-max mPDP at different problem scales of $N$ and $M$. We generate the 100 problems set with a uniform distribution of node locations per scale.
We conduct experiments with $N=200, 500, 1000, 2000, 5000$ and set $M$ such that $10\leq N/M \leq 20$ by referring practical setting of min-max routing application \cite{ma2021hierarchical}. 
% To evaluate the performance on real-world datasets, we also report results on min-max mTSP benchmark \cite{necula2015tackling} in \cref{append:mTSPLib}.

\subsubsection{Speed Evaluation.} All experiments were performed using a single NVIDIA A100 GPU and an AMD EPYC 7542 32-core processor as the CPU. Comparing the speed performance of classical algorithms (CPU-oriented) and learning algorithms (GPU-oriented) poses a significant challenge \cite{kool2018attention,kim2021learning}, given the need for a fair evaluation. While certain approaches exploit the parallelizability of learning algorithms on GPUs, enabling faster solutions to multiple problems than classical algorithms, our method follows a serial approach in line with the prior min-max learning methods \cite{park2021schedulenet, nce}. Note that when we leverage the parallelizability of our method, our approach can achieve speeds more than $100 \times$ faster; refer to Appendix A.

\subsection*{Performance Evaluation on mTSP}

\subsubsection{Baselines for mTSP.}  We consider two representative deep learning-based baseline algorithms: the ScheduleNet \citep{park2021schedulenet} and Neuro Cross Exchange \citep{nce} for min-max mTSP. For conciseness, We denote them as SN and NCE respectively. We have also included two classical heuristic methods, namely LKH3 \citep{lkh2017} and OR-Tools \citep{ortools}, with respective time limits of 60 seconds and 600 seconds per instance. Specifically, LKH3 utilizes $\lambda$-opt improvement iterations to enhance the solution within the given time budget. The time limit directly influences the number of iterations performed (following the approach in \citet{xin2021neurolkh}). Similarly, OR-Tools incorporates an iterative local search procedure for solution improvement, with the time limit governing the iterations of the local search.

\subsubsection{Results.} The results in \cref{tab:mtsp}. demonstrate that the \ourmethod{} (denoted as `ET' in tables and figures) outperforms all baselines with impressive speed. As the problem scale increases, the performance gap between ET and other methods widens further. Specifically, for $N=1000$ and $M=100$, ours achieves a cost of 2.05, significantly better than LKH3 (2.92) and NCE (2.16). Moreover, our method is $15.01/1.79 \approx 8.39\times$ faster than NCE and about $600/1.79 \approx 335\times$ faster than LKH3. The time-performance trade-off analysis shown in \cref{fig:tradeoff_mtsp} confirms that our method outperforms every baseline and provides the Pareto frontier on multi-objective of time and cost.

For the large-scale problem of $N=5000$, the ScheduleNet suffers from the complexity inherent in the parallel planning, failing to produce a solution within 10,000 seconds per problem, making it out-of-budget ($OB$). While LKH3 and OR-Tools methods can provide solutions within the allotted time, their performance is inadequate due to the inherent difficulty of large-scale problems, requiring a significantly higher number of improvement iterations for low-cost solutions. On the other hand, the NCE method surpasses classical approaches, as claimed in their main paper, but our method outperforms NCE by a substantial margin of approximately $290/8.78 \approx 33\times$ faster speed and $(2.97-2.40)/2.97 \approx 19 \%$ reduced cost.

\subsection*{Performance Evaluation on mPDP}
\begin{table*}[t!]
\begin{center}
\vspace{0.01in}
\resizebox{0.98\linewidth}{!}{
\begin{tabular}{
c 
c 
c 
c 
c
c 
c
c
c 
c
}
\toprule
&
& \multicolumn{2}{c}{Classic-based}
& \multicolumn{6}{c}{Learning-based}

\\
\cmidrule(lr{0.2em}){3-4}
\cmidrule(lr{0.2em}){5-10} 
\multicolumn{1}{c}{$N$}
& \multicolumn{1}{c}{$M$} 
& \multicolumn{1}{c}{OR-Tools (60)}
& \multicolumn{1}{c}{OR-Tools (600)}
& \multicolumn{1}{c}{AM }
& \multicolumn{1}{c}{AM$^\dagger$ }
& \multicolumn{1}{c}{HAM }
& \multicolumn{1}{c}{HAM$^\dagger$ }
& \multicolumn{1}{c}{ET (\textit{ours})}
& \multicolumn{1}{c}{ET$^\dagger$ (\textit{ours})}
\\
\midrule

\multirow{4}{*}{\shortstack[l]{200}}
&10
&20.96 \small{(60)}
&18.76 \small{(600)}
&15.88 \small{(0.33)}
&15.65 \small{(0.67)}
&5.69 \small{(0.33)}
&5.30 \small{(0.55)}
&5.03 \small{(0.54)}
&\textbf{4.68} \small{(0.55)}

\\
\addlinespace
&15
&13.96 \small{(60)}
&8.46 \small{(600)}
&15.88 \small{(0.33)}
&15.57 \small{(0.69)}
&5.21 \small{(0.34)}
&5.09 \small{(0.57)}
&3.91 \small{(0.55)}
&\textbf{3.65} \small{(0.56)}

\\
\addlinespace
&20 
&10.67 \small{(60)}
&5.70 \small{(600)}
&15.88 \small{(0.35)}
&15.55 \small{(0.71)}
&5.21 \small{(0.35)}
&5.09 \small{(0.61)}
&3.39 \small{(0.56)}
&\textbf{3.18} \small{(0.61)}

\\
\midrule
\multirow{4}{*}{\shortstack[l]{500}}
&30
&16.99 \small{(60)}
&16.99 \small{(600)}
&26.98 \small{(0.82)}
&26.15 \small{(3.10)}
&9.10 \small{(0.84)}
&8.86 \small{(1.92)}
&4.38 \small{(1.33)}
&\textbf{4.11} \small{(1.55)}

\\
\addlinespace
&40
&12.99 \small{(60)}
&12.65 \small{(600)}
&26.98 \small{(0.86)}
&26.14 \small{(3.17)}
&9.10 \small{(0.84)}
&8.87 \small{(1.95)}
&3.75 \small{(1.36)}
&\textbf{3.52} \small{(1.62)}

\\
\addlinespace
&50 
&10.99 \small{(60)}
&10.41 \small{(600)}
&26.98 \small{(0.85)}
&26.14 \small{(3.20)}
&9.10 \small{(0.88)}
&8.87 \small{(1.95)}
&3.44 \small{(1.38)}
&\textbf{3.23} \small{(1.66)}

\\
\midrule
\multirow{4}{*}{\shortstack[l]{1000}}
&50
&21.00 \small{(60)}
&21.00 \small{(600)}
&40.86 \small{(1.61)}
&39.63 \small{(11.28)}
&15.12 \small{(1.63)}
&14.58 \small{(6.35)}
&4.91 \small{(2.63)}
&\textbf{4.73} \small{(3.56)}

\\
\addlinespace
&75
&14.00 \small{(60)}
&14.00 \small{(600)}
&40.86 \small{(1.68)}
&39.61 \small{(11.44)}
&15.12 \small{(1.70)}
&14.59 \small{(6.41)}
&3.96 \small{(2.65)}
&\textbf{3.77} \small{(3.63)}

\\
\addlinespace
&100 
&11.00 \small{(60)}
&10.98 \small{(600)}
&40.86 \small{(1.69)}
&39.63 \small{(11.24)}
&15.12 \small{(1.72)}
&14.61 \small{(6.61)}
&3.56 \small{(2.75)}
&\textbf{3.38} \small{(3.80)}

\\
\midrule
\multirow{4}{*}{\shortstack[l]{2000}}
&100
&21.00 \small{(60)}
&21.00 \small{(600)}
&62.85 \small{(3.24)}
&61.31 \small{(24.98)}
&25.68 \small{(3.40)}
&25.06 \small{(15.17)}
&5.15 \small{(5.22)}
&\textbf{4.91} \small{(9.22)}

\\
\addlinespace
&150
&14.00 \small{(60)}
&14.00 \small{(600)}
&62.85 \small{(3.34)}
&61.28 \small{(25.43)}
&25.68 \small{(3.40)}
&25.04 \small{(16.26)}
&4.17 \small{(5.31)}
&\textbf{3.97} \small{(9.50)}

\\
\addlinespace
&200
&11.00 \small{(60)}
&11.00 \small{(600)}
&62.85 \small{(3.35)}
&61.33 \small{(26.33)}
&25.68 \small{(3.50)}
&25.06 \small{(16.47)}
&3.79 \small{(5.43)}
&\textbf{3.62} \small{(10.01)}

\\
\midrule
\multirow{4}{*}{\shortstack[l]{5000}}
&300
&17.00 \small{(60)}
&17.00 \small{(600)}
&114.73 \small{(8.30)}
&112.84 \small{(180)}
&54.07 \small{(34.65)}
&53.46 \small{(279)}
&4.81 \small{(52.66)}
&\textbf{4.60} \small{(79.23)}

\\
\addlinespace
&400
&13.00 \small{(60)}
&13.00 \small{(600)}
&114.73 \small{(8.31)}
&112.90 \small{(182)}
&54.07 \small{(34.44)}
&53.43 \small{(283)}
&4.33 \small{(54.86)}
&\textbf{4.11} \small{(82.59)}

\\
\addlinespace
&500
&11.00 \small{(60)}
&11.00 \small{(600)}
&114.73 \small{(8.33)}
&112.83 \small{(186)}
&54.07 \small{(34.46)}
&53.45 \small{(286)}
&4.12 \small{(54.77)}
&\textbf{3.88} \small{(82.87)}

\\
\bottomrule
\end{tabular}%
}
\end{center}
\caption{
{Results on min-max mPDP. Every performance is average performance among 100 instances. The bold symbol indicates the best performance. Average running times (in seconds) are provided in brackets. 
}
}
\label{tab:mpdp}
% \vspace{-5pt}
\end{table*}
\begin{figure*}[t!]
\centering
% \begin{subfigure}{.30\textwidth}
% \centering
% \includegraphics[width=1.0\textwidth]{figure/er400_500.png}
% \caption{mPDP with $(50, 5)$}
% \end{subfigure}
%\hspace{0.2in}
\begin{subfigure}{.40\textwidth}
\centering
\includegraphics[width=0.8\textwidth]{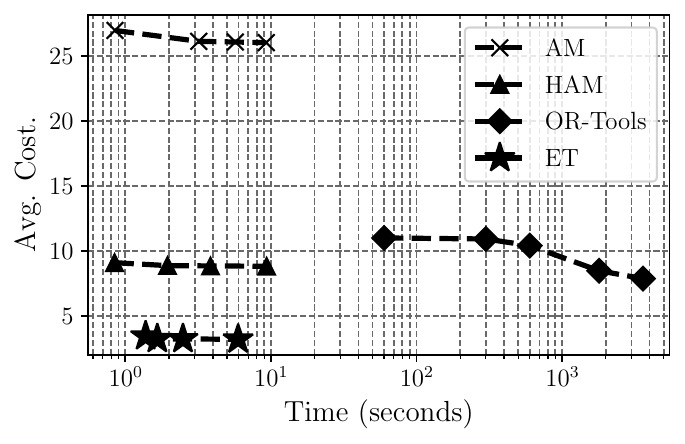}
\caption{mPDP with $(500, 50)$}
\end{subfigure}
\hspace{0.2in}
\begin{subfigure}{.40\textwidth}
\centering
\includegraphics[width=0.8\textwidth]{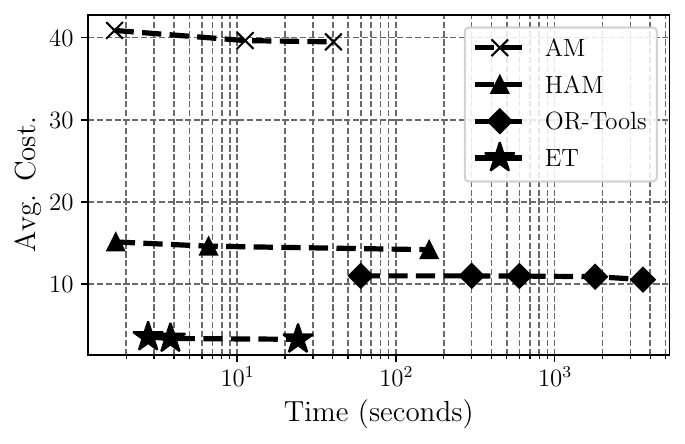}
\caption{mPDP with $(1000, 100)$}
\end{subfigure}
\caption{
Time-performance trade-off graph for mPDP.
The left and bottom indicate the Pareto frontier.}
\label{fig:tradeoff_mpdp}\vspace{-5pt}
\end{figure*}

\subsubsection{Baselines for mPDP.} We consider representative two deep learning-based baseline algorithms: AM \citep{kool2018attention} and heterogeneous AM \citep{li2021heterogeneous}, denoted as HAM. We retrain AM, and HAM using min-max objective; the details are provided in Appendix B.
%see \cref{append:exp-setting} for detailed implementation.
We also marked $\dagger$ by giving more trials for inference solutions such as sampling width \cite{kool2018attention} and augmentation width \cite{kwon2020pomo}; see Appendix B for the details.
%see \cref{append:exp-setting} for detailed setting.
We also include a heuristic method, OR-Tools \cite{ortools}, while LKH3 \cite{lkh2017} cannot handle min-max mPDP to the best of our knowledge. We exclude the multi-agent PDP (MAPDP) model \cite{zong2022mapdp} due to the inaccessible source code. 

The results presented in \cref{tab:mpdp} demonstrate that our methods (i.e., ET and ET$^\dagger$) outperform all other baselines, aligning with the findings from the mTSP experiments. Compared to OR-Tools, ET$^\dagger$ exhibits a remarkable speed improvement of $600/0.55 \approx 1901 \times$, while reducing the objective cost by approximately $(18.76-4.68)/18.76 \approx 75\%$ at $N=200, M=10$. Moreover, as shown in \Cref{fig:tradeoff_mpdp}, our method consistently presents the Pareto frontier compared to other baselines.

Importantly, in certain instances, both AM and HAM produce identical cost values as the number of agents $M$ increases. For instance, when $N=500$, AM and HAM yield the same scores for $M=30, 40, 50$. These methods were primarily designed to address min-sum problems (with HAM especially focusing on min-sum mPDP), exhibiting a limited emphasis on leveraging the concept of \textit{equity} among agents. These findings serve as compelling evidence supporting the success of our design choice centered around equity considerations.

\subsection*{Ablation Study}

To assess the influence of each component within our methodology on performance enhancement, we conducted an ablation study. As illustrated in \cref{table:ablation}, both components of our approach yielded significant performance improvements. Notably, the $\emptyset$ configuration, which represents the absence of these components, resulted in the poorest performance, indicating that a straightforward application of sequential planning to the min-max routing problem is not inherently promising. However, when we combined the multi-agent positional encoder (MPE) and the context encoder (CE), we observed substantial performance improvements, particularly in larger-scale scenarios.

% This finding suggests that our novel Transformer design for min-max routing problems significantly contributes to the performance improvement observed in our approach.

% Our results demonstrate the importance of both multi-agent positional encoding (MPE) at the encoder and the context encoder (CE) for supporting the decoder. With respect to MPE, we posit that incorporating an order bias into idle agents at the encoding stage enables the modeling of simultaneous decision-making as a sequential process.

\begin{table}[t]
\centering
% \vspace{-0.1in}
\resizebox{0.9\linewidth}{!}{
\begin{tabular}{lccccccccc}
\toprule 
\multicolumn{1}{r}{$N$}&
\multicolumn{3}{c}{100}&     
\multicolumn{3}{c}{200}    
\\
\cmidrule(lr{0.2em}){1-1}
\cmidrule(lr{0.2em}){2-4}
\cmidrule(lr{0.2em}){5-7}
\multicolumn{1}{r}{$M$}&
5
&
10
&
15
&
10
&
15
&
20
\\ \midrule
$\emptyset$                     
& 2.86          
& 2.12          
& 2.12          
& 2.92          
& 2.90  
& 2.90\\
$\{\text{MPE}\}$                
& 2.35 
& 1.97  
& 1.96          
& 2.51          
& 2.33  
& 2.80\\
$\{\text{CE}\}$               
& 2.52
& 1.97
& 1.95
& 2.28         
& 2.01
& 1.98 \\
$\{\text{MPE},\text{CE}\}$ 
& \textbf{2.35}  
& \textbf{1.96}  
& \textbf{1.95} 
& \textbf{2.15}
& \textbf{1.99} 
& \textbf{1.98}\\
% $\{\text{MPE},\text{CE},\text{CF}\}$ 
% & \textbf{2.29}  
% & \textbf{1.95}  
% & \textbf{1.95} 
% & \textbf{2.06} 
% & \textbf{1.97} 
% & \textbf{1.96} \\
\bottomrule
\end{tabular}
}
\caption{Ablation study for the combination of \ourmethod{} components. The $\emptyset$ represents the AM without our components. The MPE and CE stand for the multi-agent positional encoding and context encoder.}
% The MPE stands for the multi-agent positional encoder. The CE stands for the context encoder.}
\label{table:ablation}
\end{table}

\subsubsection{Ablation Study for Order Bias.}
\begin{figure}[t!]
\centering
% \begin{subfigure}{.30\textwidth}
% \centering
% \includegraphics[width=1.0\textwidth]{figure/er400_500.png}
% \caption{mPDP with $(50, 5)$}
% \end{subfigure}
%\hspace{0.2in}
\begin{subfigure}{.21\textwidth}
\centering
\includegraphics[width=1.0\textwidth]{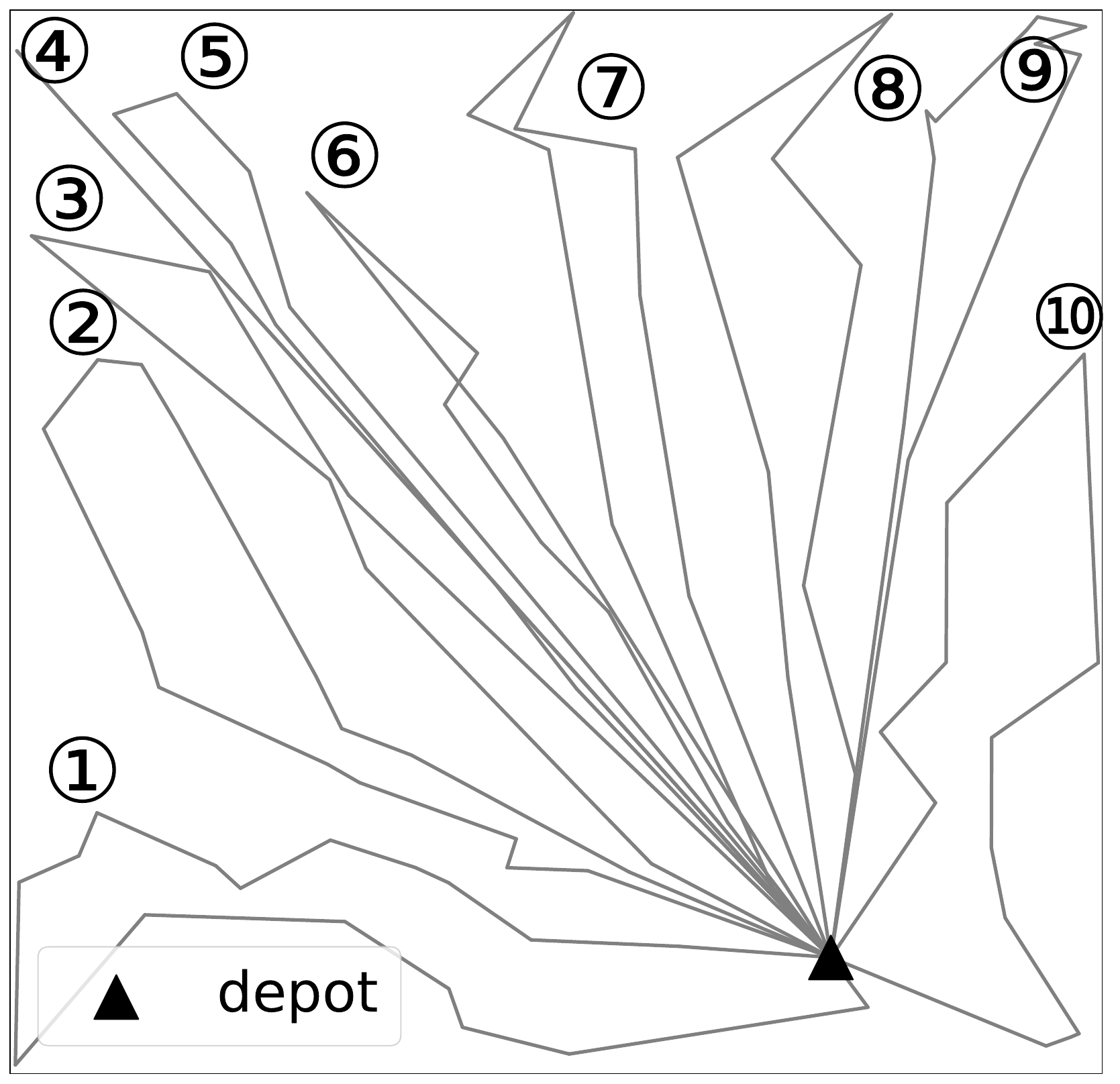}
\caption{With MPE}
\end{subfigure}
\hspace{0.2in}
\begin{subfigure}{.21\textwidth}
\centering
\includegraphics[width=1.0\textwidth]{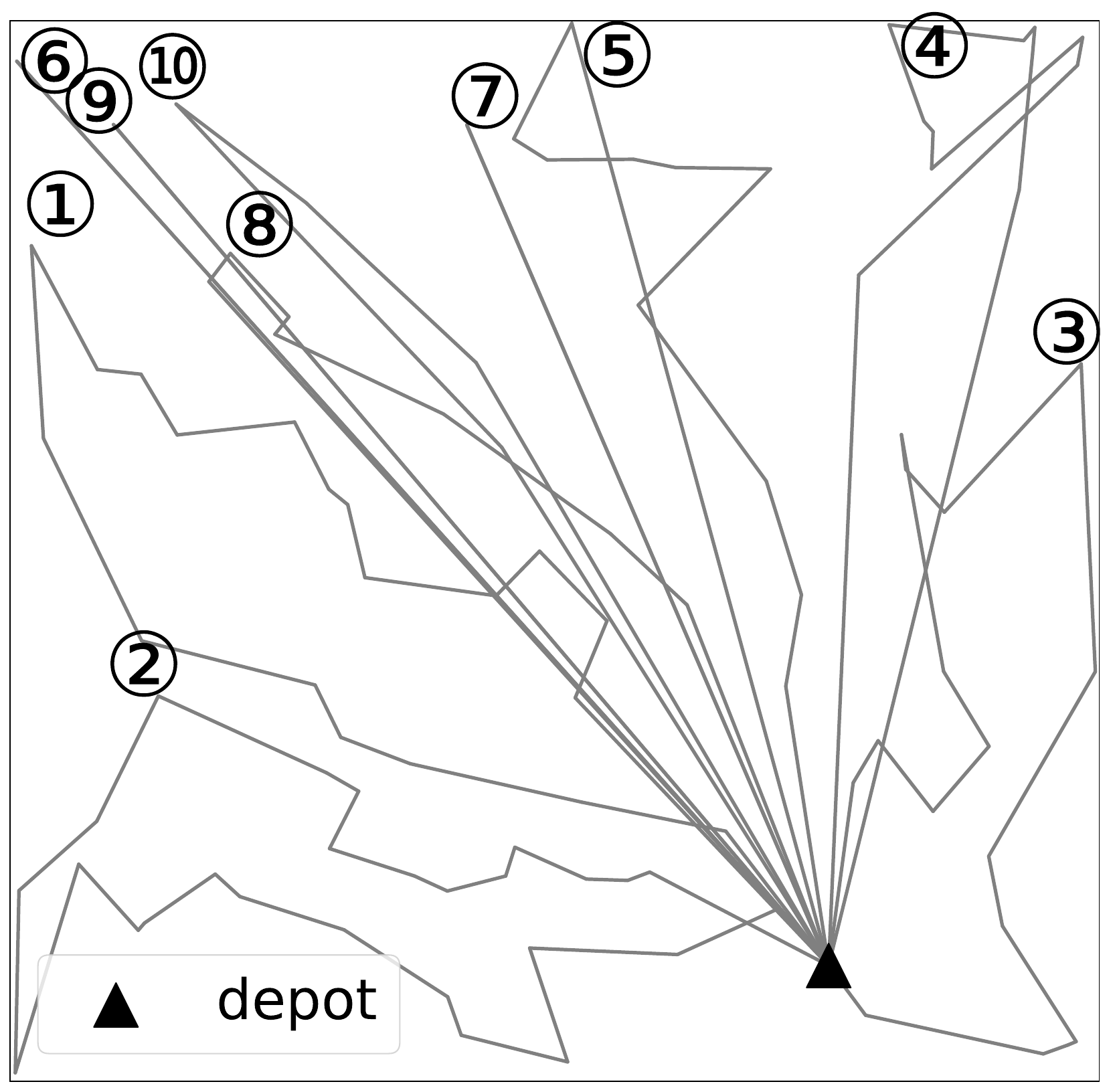}
\caption{Without MPE}
\end{subfigure}
\caption{
Ablation study for the multi-agent positional encoding (MPE) on mTSP with $N=100$ and $M = 10$.}
% \vspace{-15pt}
\label{fig:abl_mpe}
\end{figure}

As depicted in \Cref{fig:abl_mpe}, MPE contributes to inducing an order bias among agents by generating cyclic sub-tours in the Euclidean space with specific orders. This can be interpreted as successful modeling of tour generation in the Euclidean space from multiple agents in the sequence space, which is the primary objective of MPE.

% See 

% % As shown in @@TODO: fig@@ %\cref{fig:cyclic_analysis}
% % , the MPE contributes to creating an order bias among agents by generating cyclic sub-tours in the Euclidean space with specific orders. This can be interpreted as successful modeling of tour generation in the Euclidean space from multiple agents in the sequence space, which is the primary objective of MPE.

\subsection*{Additional Experiments}

% \subsubsection{Validating Performance on Real-world Benchmark of min-max mTSP.} We validate \ourmethod{} in real-world benchmarks; see \Cref{append:mTSPLib}.
% %See \Cref{append:mTSPLib}.

% \subsubsection{Validating Robustness on the Distributional Shift of the Problem.} We validate the robustness of \ourmethod{} in five problem distributions compared with LKH3; see \Cref{append:dist_shift}
% %See \Cref{append:distribtution}.

% \subsubsection{Validating Robustness on Scale shift of the $N/M$. }\Cref{append:scale_shift} We validate robustness of \ourmethod{} when the ratio of number of cities $N$ and number of agents $M$ changes comparing with LKH3. 

% \subsubsection{Comparison with the two-stage solver.} We compare \ourmethod{} with state-of-the-art two-stage solver, SplitNet \cite{liang2023splitnet}, designed for min-max mTSP; see \Cref{append:bi-level}. 
% }

We conducted several additional experiments in the Appendix C and Appendix D. First, we validate the performance of the \ourmethod{} in a real-world benchmark dataset (Appendix C). Among the baseline methods, \ourmethod{} demonstrates superior performance in almost all instances.
Moreover, we assessed the robustness of \ourmethod{} under various problem distributions (Appendix D.1) and different $N/M$ ratios (Appendix D.2), comparing it with LKH3. These experiments confirmed the robustness of \ourmethod{} to the changes in problem distributions and $N/M$ ratios.
Lastly, we compared \ourmethod{} with two competitive two-stage mTSP solvers \citep{hu2020reinforcement, liang2023splitnet}, and ours outperformed them in terms of performance (Appendix D.3).

%See \Cref{append:scale-shift}.

\section*{Related Work}
\subsection*{Vehicle Routing Problems}

After \citet{pointer} suggested the Pointer Networks, which constructively generate permutation sequences as routing solutions, termed \textit{constructive solver}, \citet{bello2017neural} turns it into deep reinforcement learning. \citet{kool2018attention} reinvent the Pointer Network using a transformer, termed attention model (AM), which becomes standard architecture for solving vehicle routing problems. By extending AM into various applications, including those outlined in recent research \cite{li2021heterogeneous,jiang2022learning,ma2021hierarchical,xin2021multi,ma2021hierarchical,ma2022efficient}, several challenges within the field of vehicle routing are addressed. In recent studies, there has been a notable emphasis on assessing the robustness of neural solvers concerning both scale shift \cite{hottung2021efficient,pmlr-v202-son23a} and distributional shift \cite{jiang2022learning,bilearning,zhou2023towards}.

Independent of constructive solution generation, other studies try to solve VRPs by learning to iteratively revise the solution, terms \textit{improvement solver}. 
\citet{chen2019learning,li2021learning,kim2021learning,wang2021rewriting} leverages local solver to rewrite partial tour to improve solution. Some studies train existing local search solvers such as 2-opt heuristic \cite{drl-2opt,wu2021learning}, large neighborhood search \cite{nlns}, iterative dynamic programming \cite{kool_dp}, and LKH \cite{xin2021neurolkh} using deep learning. Some studies use fine-tuning schemes focused on test-time adaptation in iterative learning \cite{hottung2021efficient, choo2022simulation}. While a constructive solver is invaluable for quickly generating an initial feasible solution, an improvement solver plays a crucial role in refining the solution to achieve enhanced optimality. These two approaches are fundamentally distinct and orthogonal in their objectives.

\subsection*{Min-Max Vehicle Routing Problems}

Most deep learning-based VRPs studied focus on min-sum routing which focuses on minimizing total tour length among multiple agents. The min-max routing problem focuses on minimizing the maximum tour length among multiple agents, making it highly relevant for time-critical tasks such as disaster management and vaccine delivery. The min-max routing method considers the equity of tours among the multiple agents \citep{francca1995m}. 

In constructive approaches, \citet{cao2021dan} and \citet{park2021schedulenet} advocate for a constructive solver that models decentralized parallel decisions made by multiple agents. Additionally, in addressing the specific challenge of min-max mTSP with time windows and rejections, \citet{zhang2022learning} introduce a constructive solver leveraging a graph neural network in conjunction with meticulous training and inference strategies.

On the other hand, in the domain of improvement-based methodologies, \citet{nce} propose an enhancement solver that learns to optimize tour components through cross-exchanges. Meanwhile, \citet{hu2020reinforcement} and \citet{liang2023splitnet} advocate a two-stage solver approach, wherein the initial stage employs a constructive solver, followed by an improvement solver tasked with refining the solution further.

\section*{Conclusion}

This paper introduced \ourmethod{}, sequential models for min-max routing problems. Our method outperformed the existing classic methods and state-of-the-art neural solvers, achieving a Pareto frontier in balancing cost and runtime on representative tasks like mTSP and mPDP. Our method demonstrates its scalability, handling large-scale cities with up to $N=5000$ nodes and agent fleets of up to $M=500$. \ourmethod{} holds potential for broader applications in general min-max vehicle routing problems, which we identify as a promising avenue for future research.

% We envision wide-ranging applications of \ourmethod{} in real-world services, particularly in multiple pick-up and delivery (mPDP) scenarios. Through our experiments, we have demonstrated the superior performance of \ourmethod{}, positioning it as a valuable tool for various delivery service industries aiming to enhance efficiency and reduce costs. 

% It is important to note that our current focus has been on solving single-depot routing problems which is the current limitation. As future work, we anticipate expanding \ourmethod{} to encompass multi-depot routing problems. 

\clearpage
\section*{Acknowledgements}

We thank anonymous reviewers for providing helpful feedback for preparing our manuscripts. This work was supported by a grant of the KAIST-KT joint research project through AI2XL Laboratory, Institute of convergence Technology, funded by KT [Project No. G01210696, Development of Multi-Agent Reinforcement Learning Algorithm for Efficient Operation of Complex Distributed Systems].

\bigskip

\bibliography{aaai24}

\appendix

\onecolumn
\section{Speed Evaluation for Serial and Parallel Process}\label{append:speed}
This section provides an evaluation of the speed performance of \ourmethod{} in two distinct scenarios: the serial process, where instances are solved one by one in a sequential manner, and the parallel process, where instances are solved concurrently. The results, illustrated in \cref{tab:speed-para}, clearly demonstrate that \ourmethod{} exhibits remarkable parallelization capabilities, resulting in a speed improvement of approximately 153 times compared to the serial process for the mTSP with 500 instances and $N=1000$. This significant enhancement underscores the potential of \ourmethod{} to efficiently handle multiple routing problem requests simultaneously in real-world situations. It is worth noting that even in the serial process, as shown in the main table, \ourmethod{} surpasses the baseline method in terms of speed, without leveraging any parallelization advantages.
\begin{table}[h]
    \centering
        
 \scalebox{1.0}{\begin{tabular}{llll}
    \toprule

    $N$ &Number of instances & Serial Process & Parallel Process\\
        \midrule
        \multirow{3}{*}{\shortstack[l]{200}}
         &10&  3.30s& 1.36s\\
         &100& 33.67s & 1.51s\\
         &500& 169.55s & 2.98s \\
        \midrule
    \multirow{3}{*}{\shortstack[l]{500}}
         &10&  8.37s& 1.93s\\
         &100&  84.13s& 2.84s\\
         &500&  449.05s& 3.68s \\
        \midrule
    \multirow{3}{*}{\shortstack[l]{1000}}
         &10& 16.48s & 3.02s \\
         &100& 163.75s & 4.53s \\
         &500& 815.52s & 5.35s \\

         \bottomrule
    \end{tabular}}
\label{tab:speed-para}
\caption{Comparison of total time for solving instances in mTSP with the serial process and parallel process in \ourmethod{}. The s stands for the seconds.}
\end{table}
\clearpage

\section{Detail of Experimental Setting}\label{append:exp-setting}
\subsection{Datasets}

In the mTSP experiment, we generate a random mTSP instance by randomly selecting $N$ nodes from the unit square. As per convention, we designate the depot as the first index among the $N$ nodes.
Similarly, in the mPDP experiment, we generate the locations of the depot and customer nodes (pickup and delivery pairs) randomly and independently. Note that the first half of the customer nodes are assigned as pickup nodes, while the second half serves as delivery nodes.

\subsection{\ourmethod{}}
During the training step of \ourmethod{} (ET), we adopt the same hyperparameters as described in \cite{kool2018attention}. In the finetuning step, we specifically focus on adjusting the $\theta_{\text{context}}$ parameter with $N=200,500$ for mTSP and $N=200$ for mPDP. For inference instances, we employ the finetuned model with $N=200$ for mTSP $N=200$, $N=500$ for mTSP $N>500$, and $N=200$ for mPDP. Additionally, we incorporate an augmentation technique following the methodology outlined in \citet{kwon2020pomo}. Finally, ET$^\dagger$ incorporates a sampling strategy where 100 samples are taken for each instance, and the best result among those samples represents the performance of ET$^\dagger$.

\begin{table}[ht]
    \centering

    \scalebox{1.0}{\begin{tabular}{llll}
    \toprule

    &Hyperparameters&mTSP & mPDP\\
        \midrule
        \multirow{4}{*}{\shortstack[l]{Training}}
         & Learning rate & 1e-4 & 1e-4\\
         & Batch-size & 512 & 512\\
         & Epochs & 100 & 100 \\
         & Epoch size &1,280,000 & 1,280,000\\
        \midrule
    \multirow{3}{*}{\shortstack[l]{Finetuning}}
         & Learning rate & 1e-5 & 1e-5\\
         & Batch-size & 128 & 128\\
         & Finetuning-time & 15h & 15h \\
        \midrule
    \multirow{1}{*}{\shortstack[l]{Inference}}
         & Augmentation & 8 & 8\\
         \bottomrule
    \end{tabular}}
\caption{Hyperparameter setting for \ourmethod{} in mTSP and mPDP.}
\end{table}

\subsection{LKH3}

LKH3 \cite{lkh2017} is an extension of the Lin-Kernighan algorithm, an effective local-search heuristic for addressing TSP. LKH3 exhibits the capability to tackle not only TSP but also a diverse range of constrained routing problems, by minimizing a penalty function that quantifies the degree of a constraint violation. However, LKH3 does not support solving algorithms for the mPDP, limiting its application to the mTSP in our work. We use the executable program of LKH3, which is publicly available\footnote{\url{http://webhotel4.ruc.dk/~keld/research/LKH-3}}.

\begin{table}[ht]
    \centering
    \scalebox{1.0}{\begin{tabular}{ll}
    \toprule

    Name&Value\\
        \midrule  
         Max trials & 1000\\
         Runs & 1 \\
         Seed & 3333 \\
         Time limit & \{60, 300, 600, 1800, 3600\}\\
        \bottomrule
    \end{tabular}}
\caption{Parameter setting for LKH3. It is worth noting that while we set specific values for the max trials and the runs, the algorithm predominantly terminated based on the time limit condition, especially when solving large scale problems. For the remaining parameters not mentioned, we use the default values provided by the program.}
\end{table}

\clearpage
\subsection{OR-Tools}

OR-tools \cite{ortools} is an open-source software that aims to solve various combinatorial optimization problems. OR-tools offers a wide range of solvers for tackling problems such as linear programming, mixed-integer programming, constraint programming, routing problems, and scheduling. Within the realm of routing problems, OR-Tools can easily handle both mTSP and mPDP by simply incorporating additional constraints tailored to each problem. The installation guide for python OR-tools library and example codes for various problems, including mTSP and mPDP, are readily available in their official website\footnote{\url{https://developers.google.com/optimization}}.

\begin{table}[ht]
    \centering

    \scalebox{1.0}{\begin{tabular}{ll}
    \toprule

    Name&Value\\
        \midrule  
         Global span cost coefficient & 10000\\
         First solution strategy & PATH\_CHEAPEST\_ARC \\
         Local search metaheuristic & GUIDED\_LOCAL\_SEARCH \\
         Time limit & \{60, 300, 600, 1800, 3600\}\\
        \bottomrule
    \end{tabular}}
\caption{Parameter setting for OR-tools. We use the same parameters for both mTSP and mPDP. Note that the global span cost coefficient should be sufficiently large to guarantee the desired objective for the min-max tasks. For the remaining parameters not mentioned, we use the default values suggested by OR-tools' official guides.}
\end{table}

\subsection{ScheduleNet}
ScheduleNet (SN) \cite{park2021schedulenet} is a method that utilizes a graph neural network (GNN) to sequentially generate simultaneous actions for multiple agents, effectively capturing the relationships between them. To conduct our experiment, we obtained the source code by contacting the author and  followed the training procedure outlined in \cite{park2021schedulenet}.

\begin{table}[ht]
    \centering
    \scalebox{1.0}{\begin{tabular}{ll}
    \toprule

    Name&Value\\
        \midrule  
         Learning rate & 1e-4\\
         Batch-size & 512 \\
         Epochs & 10,000 \\
         Epoch size &65,536\\
         Discounting factor & 0.9\\
         Smoothing coefficient & 0.1\\
         Clipping parameter & 0.2\\
        \bottomrule
    \end{tabular}}
\caption{Hyperparameter for training ScheduleNet in mTSP.}
\end{table}

\subsection{Neuro Cross Exchange}
The Neuro Cross Exchange (NCE) \cite{nce} is a supervised-learning-based improvement solver and a neural meta-heuristic technique that addresses vehicle routing problems (VRPs) by strategically swapping sub-tours among the vehicles to enhance solutions. We train a cost-decrement prediction model ($f_\theta(\cdot)$) using a number of cities ($N_c$) sampled from a uniform distribution $U(50, 100)$. The number of depots ($N_d$) is set to 1, and the number of vehicles ($N_v$) is 2. The 2D coordinates of the cities are sampled from $U(0, 1)$. With $N_v = 2$, we generate two tours using the greedy assignment heuristic. From these two tours ($\tau_1$, $\tau_2$), we compute the best true cost-decrements for all feasible combinations to create the training dataset. Overall, we generated 47,856,986 training samples from 50,000 instances. The model $f_\theta(\cdot)$ is parameterized using a Graph Neural Network (GNN) with five layers of the attentive embedding layer. Additionally, we employ four-layered Multi-Layer Perceptrons (MLPs) to parameterize $\phi_e$, $\phi_w$, $\phi_n$, and $\phi_c$, with hidden dimensions of 64 and the Mish activation function.

We obtained the source code directly from the author by email. And we restrict the time because the improvement method will get the optimal solution when it has enough time. \Cref{tab:ncetime} describes the time limits we set for each problem size.

\begin{table}[ht]
    \centering
      
    \scalebox{1.0}{\begin{tabular}{ll}
    \toprule
    Number of cities ($N$) &Time limit\\
        \midrule
         200 & 5.00\\
         500& 5.00\\
         1,000 & 15.00\\
         2,000& 30.00\\
         5,000 & 180.00\\
        \bottomrule
    \end{tabular}}
    
\caption{Time limit for NCE in mTSP.}
\label{tab:ncetime}
\end{table}

\subsection{Attention Model}

The Attention Model (AM) \cite{kool2018attention} serves as a fundamental component in numerous models, employing attention mechanisms to address a wide range of routing problems, including TSP, CVRP, PDP, and others. During the training phase, we adhere to the recommended hyperparameters provided by the open-source code\footnote{\url{https://github.com/wouterkool/attention-learn-to-route}}. Additionally, we adopt a similar strategy for fine-tuning as ET. During the inference step, AM$^\dagger$ also utilizes the same strategy as ET$^\dagger$.

\begin{table}[ht]
    \centering

   \scalebox{1.0}{\begin{tabular}{lll}
    \toprule

    &Name& Value\\
        \midrule
        \multirow{4}{*}{\shortstack[l]{Training}}
         & Learning rate & 1e-4\\
         & Bats far and the ch-size & 512\\
         & Epochs & 100 \\
         & Epoch size &1,280,000\\
        \midrule
    \multirow{3}{*}{\shortstack[l]{Finetuning}}
         & Learning rate & 1e-5\\
         & Batch-size & 128\\
         & Finetuning-time & 15h \\
        \midrule
    \multirow{1}{*}{\shortstack[l]{Inference}}
         & Augmentation & 8 \\
         \bottomrule
    \end{tabular}}
\caption{Hyperparameter setting for AM in mPDP.}
\label{tab:amhparams}
\end{table}

\subsection{Heterogeneous Attention Model}
The Heterogeneous Attention Model (HAM) \cite{li2021heterogeneous} is an adapted variation of the AM, specifically customized for addressing PDP. This model introduces additional attention mechanisms that are specifically designed to address the considerations of precedence constraint between the pickup node and the delivery node. During the training phase, we adhere to the recommended hyperparameters provided by the open-source code\footnote{\url{https://github.com/Demon0312/Heterogeneous-Attentions-PDP-DRL}}. Additionally, we adopt a similar strategy for fine-tuning as ET. During the inference step, HAM$^\dagger$ also utilizes the same strategy as ET$^\dagger$.

\begin{table}[ht]
    \centering

    \scalebox{1.0}{\begin{tabular}{lll}
    \toprule

    &Name & Value\\
        \midrule
        \multirow{4}{*}{\shortstack[l]{Training}}
         & Learning rate & 1e-4\\
         & Batch-size & 512\\
         & Epochs & 800 \\
         & Epoch size &1,280,000\\
        \midrule
    \multirow{3}{*}{\shortstack[l]{Finetuning}}
         & Learning rate & 1e-5\\
         & Batch-size & 128\\
         & Finetuning-time & 15h \\
        \midrule
    \multirow{1}{*}{\shortstack[l]{Inference}}
         & Augmentation & 8 \\
         \bottomrule
    \end{tabular}}
\caption{Hyperparameter setting for HAM in mPDP.}
\label{tab:hamhparams}
\end{table}
\clearpage

\section{Experimental Results on Real-world mTSP Dataset}\label{append:mTSPLib}
This section presents the experimental results of a real-world mTSP dataset by converting the TSPLIB \cite{reinelt1991tsplib} into mTSP instances. Following the established convention \cite{necula2015tackling} for converting TSPLIB to mTSP instances, we set the first element in the list of cities as a depot. Additionally, we set the number of agents $M$ depending on the problem size as follows: $M={10,15,20}$ for problems with $200<N<500$, $M={30,40,50}$ for problems with $500<N<1000$, and $M={50,75,100}$ for problems with $N>1000$. The results presented in \cref{tab:mtsplib} demonstrate that our ET$^\dagger$ consistently outperforms all baselines across nearly all tasks.
\begin{table*}[!ht]
\begin{center}

\resizebox{0.9\linewidth}{!}{
\begin{tabular}{
c 
c
c  
c 
c 
c 
c 
c 
c
}
\toprule
&
& \multicolumn{2}{c}{Classic-based}
& \multicolumn{4}{c}{Learning-based}
%& \multicolumn{1}{c}{Ours}
\\
\cmidrule(lr{0.2em}){3-4}
\cmidrule(lr{0.2em}){5-8} 
% \cmidrule(lr{0.2em}){9-9}
\multicolumn{1}{c}{TSPLib}
& \multicolumn{1}{c}{$M$} 
& \multicolumn{1}{c}{LKH3}
% & \multicolumn{1}{c}{LKH3 (10)}
& \multicolumn{1}{c}{OR-tools}
% & \multicolumn{1}{c}{OR-tools (10)}
& \multicolumn{1}{c}{SN}
& \multicolumn{1}{c}{NCE}
% & \multicolumn{1}{c}{MF (N.C.)}
& \multicolumn{1}{c}{ET (\textit{ours})}
& \multicolumn{1}{c}{ET$^\dagger$ (\textit{ours})}
\\
\midrule

\multirow{3.5}{*}{\shortstack[l]{kroA200}}
&10
% &2.52 \small{(102)}
% &- \small{(-)}
&6417.19 \small{(600)}
&\textbf{6223.22} \small{(600)}
&8339.22 \small{(11.72)}
&6281.18 \small{(5.02)}
% &2.15 \small{(0.03)}
&6427.29 \small{(0.35)}
&6294.89 \small{(0.70)}

\\
\addlinespace
&15
% &2.39 \small{(102)}
% &- \small{(-)}
&6417.19 \small{(600)}
&\textbf{6223.22} \small{(600)}
&6844.31 \small{(10.87)}
&6280.73 \small{(5.06)}
% &1.99 \small{(0.03)}
&\textbf{6223.22} \small{(0.35)}
&\textbf{6223.22} \small{(0.75)}

\\
\addlinespace
&20
% &2.29 \small{(102)}
% &- \small{(-)}
&6418.51 \small{(600)}
&\textbf{6223.22} \small{(600)}
&7130.81 \small{(9.88)}
&\textbf{6223.22} \small{(5.02)}
&\textbf{6223.22} \small{(0.36)}
&\textbf{6223.22} \small{(0.72)}

\\
\midrule
% \multirow{3.5}{*}{\shortstack[l]{a280}}
% &10
% % &3.31 \small{(102)}
% % &2.70 \small{(1008)}
% &648.89 \small{(600)}
% &944.42 \small{(600)}
% &715.95 \small{(27.84)}
% &643.59 \small{(5.03)}
% % &2.40 \small{(0.05)}
% &646.89 \small{(0.55)}
% &\textbf{636.00} \small{(1.25)}
% \\
% \addlinespace
% &15
% % &3.10 \small{(102)}
% % &2.55 \small{(1008)}
% &668.89 \small{(600)}
% &951.98 \small{(600)}
% &648.98 \small{(26.91)}
% &617.67 \small{(5.01)}
% &611.80 \small{(0.55)}
% &\textbf{607.78} \small{(1.35)}

% \\
% \addlinespace
% &20 
% % &2.93 \small{(1002)}
% % &2.48 \small{(1008)}
% &656.11 \small{(600)}
% &943.60 \small{(600)}
% &649.96 \small{(27.04)}
% &\textbf{604.68} \small{(5.03)}
% % &2.14 \small{(0.05)}
% &607.90 \small{(0.58)}
% &\textbf{604.68} \small{(1.30)}
% \\
% \midrule
\multirow{3.5}{*}{\shortstack[l]{lin318}}
&10
% &4.45 \small{(102)}
% &3.77 \small{(1005)}
&10296.33 \small{(600)}
&17546.77 \small{(600)}
&10842.31 \small{(29.35)}
&10042.13 \small{(5.04)}
% &3.03 \small{(0.13)}
&10023.11 \small{(0.70)}
&\textbf{9945.76} \small{(1.58)}

\\
\addlinespace
&15
% &3.71 \small{(102)}
% &3.26 \small{(1006)}
&10373.49 \small{(600)}
&18406.43 \small{(600)}
&9876.20 \small{(31.07)}
&\textbf{9731.17} \small{(5.07)}
% &2.39 \small{(0.14)}
&\textbf{9731.17} \small{(0.69)}
&\textbf{9731.17} \small{(1.60)}

\\
\addlinespace
&20
% &3.23 \small{(103)}
% &2.92 \small{(1007)}
&10854.49 \small{(600)}
&17628.86 \small{(600)}
&9933.23 \small{(34.55)}
&\textbf{9731.17} \small{(5.12)}
% &2.30 \small{(0.14)}
&\textbf{9731.17} \small{(0.64)}
&\textbf{9731.17} \small{(1.64)}

\\
\midrule
\multirow{3.5}{*}{\shortstack[l]{pr439}}
&10
% &- \small{(-)}
% &- \small{(-)}
&26206.85 \small{(600)}
&58355.61 \small{(600)}
&25807.61\small{(105)}
&29685.56 \small{(5.04)}
&25651.63 \small{(0.71)}
&\textbf{24374.03} \small{(2.37)}

\\
\addlinespace
&15
% &- \small{(-)}
% &- \small{(-)}
&22457.51 \small{(600)}
&58355.50 \small{(600)}
&22968.29 \small{(102)}
&24689.55 \small{(5.04)}
% &2.22 \small{(0.05)}
&22325.89 \small{(0.78)}
&\textbf{21833.94} \small{(2.42)}

\\
\addlinespace
&20
% &- \small{(-)}
% &- \small{(-)}
&24503.16 \small{(600)}
&58355.61 \small{(600)}
&22539.62.29 \small{(103)}
&21828.70 \small{(5.00)}
&22112.27 \small{(0.76)}
% &2.14 \small{(0.05)}
&\textbf{21703.51} \small{(2.45)}

\\
\midrule

\multirow{3.5}{*}{\shortstack[l]{u574}}
&30
% &4.45 \small{(102)}
% &3.77 \small{(1005)}
&8799.53 \small{(600)}
&19391.32 \small{(600)}
&6875.21 \small{(244)}
&\textbf{6641.51} \small{(5.02)}
&\textbf{6641.51} \small{(0.95)}
% &3.03 \small{(0.13)}
&\textbf{6641.51} \small{(1.62)}

\\
\addlinespace
&40
% &3.71 \small{(102)}
% &3.26 \small{(1006)}
&8051.49 \small{(600)}
&15923.89 \small{(600)}
&6731.26 \small{(257)}
&\textbf{6641.51} \small{(5.08)}
&\textbf{6641.51} \small{(1.03)}
% &2.39 \small{(0.14)}
&\textbf{6641.51} \small{(1.64)}

\\
\addlinespace
&50
% &3.23 \small{(103)}
% &2.92 \small{(1007)}
&7733.13 \small{(600)}
&14191.82 \small{(600)}
&6859.72 \small{(289.65)}
&\textbf{6641.51} \small{(5.16)}
& \textbf{6641.51} \small{(1.01)}
% &2.30 \small{(0.14)}
&\textbf{6641.51} \small{(1.71)}
\\
\midrule

\multirow{3.5}{*}{\shortstack[l]{p654}}
&30
% &4.45 \small{(102)}
% &3.77 \small{(1005)}
&13317.00 \small{(600)}
&25551.51 \small{(600)}
&14649.50 \small{(453.36)}
&16069.91 \small{(5.05)}
&15905.85 \small{(1.10)}
% &3.03 \small{(0.13)}
&\textbf{12794.86} \small{(1.92)}

\\
\addlinespace
&40
% &3.71 \small{(102)}
% &3.26 \small{(1006)}
&13667.81 \small{(600)}
&25547.37 \small{(600)}
&14627.26 \small{(393)}
&16246.51 \small{(5.01)}
&13016.20 \small{(1.13)}
% &2.39 \small{(0.14)}
&\textbf{12747.33} \small{(2.05)}

\\
\addlinespace
&50
% &3.23 \small{(103)}
% &2.92 \small{(1007)}
&13187.50 \small{(600)}
&25547.37 \small{(600)}
&14531.31 \small{(494.67)}
&14832.95 \small{(5.05)}
&12788.09 \small{(1.14)}
% &2.30 \small{(0.14)}
&\textbf{12501.64} \small{(2.05)}
\\
\midrule

\multirow{3.5}{*}{\shortstack[l]{rat783}}
&30
% &4.45 \small{(102)}
% &3.77 \small{(1005)}
&2217.40 \small{(600)}
&5105.11 \small{(600)}
&1380.92 \small{(628)}
&1381.56 \small{(6.26)}
&1319.97 \small{(1.30)}
% &3.03 \small{(0.13)}
&\textbf{1271.52} \small{(2.61)}

\\
\addlinespace
&40
% &3.71 \small{(102)}
% &3.26 \small{(1006)}
&1872.14 \small{(600)}
&5105.11 \small{(600)}
&1352.61 \small{(792)}
&1423.14 \small{(6.31)}
&1254.60 \small{(1.31)}
% &2.39 \small{(0.14)}
&\textbf{1237.64} \small{(2.60)}

\\
\addlinespace
&50
% &3.23 \small{(103)}
% &2.92 \small{(1007)}
&1639.60 \small{(600)}
&4004.67 \small{(600)}
&1323.87 \small{(787)}

&1410.63 \small{(6.38)}
&1249.93 \small{(1.30)}
% &2.30 \small{(0.14)}
&\textbf{1231.69} \small{(2.67)}
\\
\midrule

\multirow{3.5}{*}{\shortstack[l]{pr1002}}
&50
% &4.45 \small{(102)}
% &3.77 \small{(1005)}
&54569.48 \small{(600)}
&159502.40 \small{(600)}
&37844.23 \small{(1647)}
&34894.64 \small{(15.03)}
&\textbf{34365.02} \small{(1.64)}
% &3.03 \small{(0.13)}
&34465.98 \small{(2.02)}

\\
\addlinespace
&75
% &3.71 \small{(102)}
% &3.26 \small{(1006)}
&50657.38 \small{(600)}
&117172.89 \small{(600)}
&36793.27 \small{(1678)}
&\textbf{33861.63} \small{(15.03)}
&34263.05 \small{(1.70)}
% &2.39 \small{(0.14)}
&\textbf{33861.63} \small{(2.06)}

\\
\addlinespace
&100
% &3.23 \small{(103)}
% &2.92 \small{(1007)}
&42902.62 \small{(600)}
&135228.92 \small{(600)}
&35654.55 \small{(1855)}
&\textbf{33861.63} \small{(15.09)}
&\textbf{33861.63} \small{(1.85)}
% &2.30 \small{(0.14)}
&\textbf{33861.63} \small{(2.19)}
\\
\midrule

\multirow{3.5}{*}{\shortstack[l]{pcb1173}}
&50
% &4.45 \small{(102)}
% &3.77 \small{(1005)}
&8530.43 \small{(600)}
&35366.75 \small{(600)}
&6715.26 \small{(2530)}
&6667.08 \small{(15.01)}
&6623.27 \small{(1.93)}
% &3.03 \small{(0.13)}
&\textbf{6607.14} \small{(2.39)}

\\
\addlinespace
&75
% &3.71 \small{(102)}
% &3.26 \small{(1006)}
&9844.05 \small{(600)}
&25881.61 \small{(600)}
&6769.86 \small{(2622)}
&6539.95 \small{(15.01)}
&6562.89 \small{(1.98)}
% &2.39 \small{(0.14)}
&\textbf{6528.86} \small{(2.51)}

\\
\addlinespace
&100
% &3.23 \small{(103)}
% &2.92 \small{(1007)}
&8770.60 \small{(600)}
&25237.96 \small{(600)}
&6765.07 \small{(2392)}
&\textbf{6528.86} \small{(15.01)}
&\textbf{6528.86} \small{(2.03)}
% &2.30 \small{(0.14)}
&\textbf{6528.86} \small{(2.49)}
\\
\midrule

\multirow{3.5}{*}{\shortstack[l]{d1291}}
&50
% &4.45 \small{(102)}
% &3.77 \small{(1005)}
&11268.07 \small{(600)}
&35976.42 \small{(600)}
&11085.14 \small{(2735)}
&11163.11 \small{(16.81)}
&9955.28 \small{(2.20)}
% &3.03 \small{(0.13)}
&\textbf{9858.99} \small{(2.67)}

\\
\addlinespace
&75
% &3.71 \small{(102)}
% &3.26 \small{(1006)}
&11455.59 \small{(600)}
&35976.42 \small{(600)}
&10776.30 \small{(3242)}
&11003.23 \small{(16.74)}
&\textbf{9858.99} \small{(2.17)}
% &2.39 \small{(0.14)}
&\textbf{9858.99} \small{(2.72)}

\\
\addlinespace
&100
% &3.23 \small{(103)}
% &2.92 \small{(1007)}
&9998.52 \small{(600)}
&30250.32 \small{(600)}
&10419.44 \small{(3511)}
&10904.31 \small{(16.69)}
&\textbf{9858.99} \small{(2.21)}
% &2.30 \small{(0.14)}
&\textbf{9858.99} \small{(2.86)}
\\
\midrule

% \multirow{3.5}{*}{\shortstack[l]{rl1304}}
% &50
% % &4.45 \small{(102)}
% % &3.77 \small{(1005)}
% &50916.48 \small{(600)}
% &179168.59 \small{(600)}
% &\textit{OB}
% &45209.16 \small{(17.48)}
% &33617.21 \small{(2.11)}
% % &3.03 \small{(0.13)}
% &\textbf{33072.96} \small{(2.63)}

% \\

% \addlinespace
% &75
% % &3.71 \small{(102)}
% % &3.26 \small{(1006)}
% &39774.59 \small{(600)}
% &178270.00 \small{(600)}
% &\textit{OB}
% &42899.00 \small{(17.77)}
% &33009.00 \small{(2.80)}
% % &2.39 \small{(0.14)}
% &\textbf{32473.34} \small{(2.80)}

% \\
% \addlinespace
% &100
% % &3.23 \small{(103)}
% % &2.92 \small{(1007)}
% &53956.65 \small{(600)}
% &118561.04 \small{(600)}
% &\textit{OB}
% &41927.05 \small{(17.57)}
% &32557.78 \small{(2.26)}
% % &2.30 \small{(0.14)}
% &\textbf{32134.47} \small{(2.80)}
% \\
% \midrule

% \multirow{3.5}{*}{\shortstack[l]{u1432}}
% &50
% % &- \small{(-)}
% % &- \small{(-)}
% &39189.72 \small{(600)}
% &85050.22 \small{(600)}
% &\textit{OB}
% &15282.77 \small{(20.47)}
% &13027.50 \small{(2.31)}
% % &3.03 \small{(0.13)}
% &\textbf{12830.64} \small{(2.95)}

% \\
% \addlinespace
% &75
% % &- \small{(-)}
% % &- \small{(-)}
% &24844.31 \small{(600)}
% &80893.17 \small{(600)}
% &\textit{OB}
% &15007.52 \small{(20.54)}
% &12835.57 \small{(2.43)}
% % &2.39 \small{(0.14)}
% &\textbf{12625.89} \small{(3.08)}

% \\
% \addlinespace
% &100
% % &- \small{(-)}
% % &- \small{(-)}
% &22187.92 \small{(600)}
% &70304.41 \small{(600)}
% &\textit{OB}
% &15241.95 \small{(20.45)}
% &12521.75 \small{(2.47)}
% % &2.30 \small{(0.14)}
% &\textbf{12155.28} \small{(3.10)}

% \\
% \midrule

\multirow{3.5}{*}{\shortstack[l]{fl1577}}
&50
% &4.45 \small{(102)}
% &3.77 \small{(1005)}
&5631.54 \small{(600)}
&15232.65 \small{(600)}
&\textit{OB}
&6586.44 \small{(24.92)}
&4531.97 \small{(2.69)}
% &3.03 \small{(0.13)}
&\textbf{4158.05} \small{(3.41)}

\\
\addlinespace
&75
% &3.71 \small{(102)}
% &3.26 \small{(1006)}
&9446.27 \small{(600)}
&15232.65 \small{(600)}
&\textit{OB}
&6356.05 \small{(24.45)}
&4278.85 \small{(2.62)}
% &2.39 \small{(0.14)}
&\textbf{4071.33} \small{(3.44)}

\\
\addlinespace
&100
% &3.23 \small{(103)}
% &2.92 \small{(1007)}
&7356.55 \small{(600)}
&15232.65 \small{(600)}
&\textit{OB}
&6089.94 \small{(24.59)}
&\textbf{4069.46} \small{(2.66)}
% &2.30 \small{(0.14)}
&4087.09 \small{(3.62)}

\\
\midrule

\multirow{3.5}{*}{\shortstack[l]{u1817}}
&50
% &4.45 \small{(102)}
% &3.77 \small{(1005)}
&16776.31 \small{(600)}
&37910.67 \small{(600)}
&\textit{OB}
&7672.36 \small{(32.66)}
&6549.83 \small{(2.96)}
% &3.03 \small{(0.13)}
&\textbf{6490.56} \small{(4.13)}

\\
\addlinespace
&75
% &3.71 \small{(102)}
% &3.26 \small{(1006)}
&15685.26 \small{(600)}
&37910.67 \small{(600)}
&\textit{OB}
&7283.72 \small{(32.84)}
&6540.25 \small{(2.98)}
% &2.39 \small{(0.14)}
&\textbf{6424.96} \small{(4.07)}

\\
\addlinespace
&100
% &3.23 \small{(103)}
% &2.92 \small{(1007)}
&14071.51 \small{(600)}
&28891.86 \small{(600)}
&\textit{OB}
&7174.80 \small{(32.67)}
&6431.92 \small{(3.04)}
% &2.30 \small{(0.14)}
&\textbf{6413.51} \small{(4.26)}
\\

\bottomrule
\end{tabular}%
}
\caption{
Performance evaluation results of real-world mTSP data converted from the TSPLIB \cite{reinelt1991tsplib} benchmark dataset. The running times (in seconds) are provided in brackets. ET$^\dagger$ utilizes additional augmentation following the approach proposed in \cite{kim2022sym}. Specifically, 2000 augmentations are applied for cities with $N < 500$, 500 for cities with $500 \leq N < 500$, and 100 for cities with $1000 \leq N < 2000$. The ``OB" represents ``out-of-budget," indicating that the allocated budget for each instance is limited to 1 hour.}
\vspace{0.01in}
\label{tab:mtsplib}
\vspace{-15pt}
\end{center}
\end{table*}

\clearpage

\section{Additional Experiments} \label{append:additional}

\subsection{Robustness Experiments on the Distributional Shift of
the Problem}\label{append:dist_shift}
    This section provides the generalization capability of ET. For training, we only use instances from uniform distribution. And for testing, we use explosion, rotation, cluster, and mixed distribution as unseen distributions. We followed the approach presented in \citet{zhou2023towards} to generate an explosion and rotation distribution dataset. Additionally, the cluster and mixed data were obtained from the work of \citet{bilearning}. 

    We generated a total of 100 instances for each distinct distribution. \Cref{tab:comparison} shows that ET exhibits robustness when faced with out-of-distribution data.

\begin{table}[!ht]
\centering

\begin{tabular}{lccccc}
\toprule
 & Uniform & Explosion & Rotation & Cluster & Mixed  \\
\midrule
LKH (600s) & 2.48 (600)&2.05 (600) & 2.01 (600) & 1.41 (600) & 2.24 (600) \\
ET & \textbf{2.01 (0.92)} &\textbf{1.75 (0.90)} & \textbf{1.70 (0.91)} & \textbf{1.25 (0.90)} & \textbf{1.91 (0.90)}  \\
\bottomrule
\end{tabular}
\caption{Performance of ET and LKH across different distributions in mTSP of $N = 500, M=50$. The term `performance (second)' symbolizes both performance and time, measured in seconds.}
\label{tab:comparison}
\end{table}

\subsection{Robustness Experiments on Scale shift of the $N/M$}\label{append:scale_shift}
This section delves into an examination of the robustness to the changes in $N/M$ ratio, which signifies the average number of tasks a single agent can handle. As our main experiments show the effectiveness of ET when $10 \leq N/M \leq 20$, here we conduct experiments with much larger $N/M$ ratio, i.e., there is an abundance of tasks but a limited pool of agent resources. Specifically, we conduct experiments on mTSP using values of $N$ set at 1000, 2000, and 5000, and $M$ at 10, 20, and 20, respectively. Each of our test datasets consists of 100 instances from uniform distribution. The outcomes, as illustrated in Table \ref{tab:scale_shift}, show that ET consistently outperforms other baseline models, achieving the Pareto frontier.

\begin{table}[ht]
\centering

\vspace{0.1cm}
\begin{tabular}{lcccccc}
\toprule
&\multicolumn{2}{c}{$N = 1000$, $M=10$}& \multicolumn{2}{c}{$N = 2000$, $M=20$} & \multicolumn{2}{c}{$N = 5000$, $M=20$}\\
\cmidrule(lr{0.2em}){2-3}\cmidrule(lr{0.2em}){4-5}\cmidrule(lr{0.2em}){6-7}
  & Performance & Time & Performance & Time & Performance & Time \\
\midrule
LKH & 5.20 & 600s & 7.52 & 600s & 13.88 & 3600s \\
OR-Tools & 13.98 & 600s & 22.28 & 600s & 35.22 & 3600s\\
ET  & \textbf{4.65} & \textbf{2.01s} & \textbf{4.64} & \textbf{3.80s} & \textbf{10.48} & \textbf{9.44s} \\

\toprule
\end{tabular}
\caption{Performance evaluation of ET and other baselines on mTSP with larger $N/M$ ratio.}
\label{tab:scale_shift}
\end{table}

\clearpage

\subsection{Comparison with two-stage solvers}\label{append:bi-level}

In this section, we compare various two-stage solvers, GNN-DisPN \cite{hu2020reinforcement} and SplitNet \cite{liang2023splitnet}. GNN-DisPN assigns tasks to agents, and then each agent constructs a tour using a standard optimization algorithm. SplitNet is the latest state-of-the-art two-stage min-max mTSP solver. SplitNet solves an mTSP problem by iteratively splitting and reorganizing edges. To ensure a fair comparison, we incorporate ET into a two-stage approach. We use ET to obtain an initial solution, and we improve it using NCE \cite{nce}, which re-optimize a given solution by swapping tours of agents. We refer to this strategy as ET*. Since there is no publicly accessible source code for SplitNet, we take the result of both Splitnet and GNN-DisPN from the SplitNet paper. 

We then evaluated our method on each of these datasets. As shown in \Cref{tab:bi-level} ET* exhibits significantly faster and superior performance compared to other baseline methods. This highlights ET's potential to effectively support bi-level approaches by delivering high-quality initial solutions more rapidly.

\begin{table}[h]
\centering

\vspace{0.1cm}
\begin{tabular}{lcccccc}
\toprule
&\multicolumn{2}{c}{$N = 200$, $M=10$}& \multicolumn{2}{c}{$N = 400$, $M=10$}  & \multicolumn{2}{c}{$N = 1000$, $M=10$}\\
\cmidrule(lr{0.2em}){2-3}\cmidrule(lr{0.2em}){4-5}\cmidrule(lr{0.2em}){6-7}
  & Performance & Time & Performance & Time & Performance & Time \\
\midrule
GNN-DisPN & 2.97 &- &7.75 &- &14.63 &-\\
SplitNet (s.64) & 2.27 & 8.6s & 2.59 & 14.3s & 3.32 & 55.1s \\
ET*  &  \textbf{2.05}&  \textbf{0.36s}&  \textbf{2.50}&  \textbf{5.12s}& \textbf{3.27} & \textbf{25.27s} \\

\toprule
\end{tabular}
\caption{Performance evaluation of ET and two-stage solvers on mTSP}
\label{tab:bi-level}
\end{table}

\subsection{Experimental Result on mCVRP}\label{append:mcvrp}
In this section, we present experimental results for the min-max Capacitated Vehicle Routing Problem (CVRP). The min-max CVRP is one of the min-max VRP problems which is composed of nodes representing customer locations and a depot serving as the starting point for vehicles. Each customer has a specified demand that our vehicle must fulfill while minimizing the maximum route length among all vehicles. Each customer is exclusively attended to by a single vehicle, and the vehicle routes are planned according to their respective load-carrying capacity constraints.
 
We bring the table from NCE paper \citep{nce}. We follow \citet{bogyrbayeva2023deep} to generate test dataset with 100 random instances. The result in \Cref{tab:cvrp} illustrates the adaptability of ET in tackling various min-max routing problems, establishing it as a versatile framework for addressing such challenges.

\begin{table}[ht]
\centering

\vspace{0.1cm}
\begin{tabular}{lcc}
\toprule
&\multicolumn{2}{c}{$N = 30$, $M=3$} \\
\midrule
  & Performance & Time \\
\midrule
OR-Tools & 2.44 & 1.0s\\
AM & 2.47 & 0.2s \\
AM (s.1200) & 2.29 & 27.6s \\
HM & 2.39 & 0.2s \\
HM (s.1200) & 2.27 & 25.2s \\
NCE & 2.25 & 2.03s \\
ET & \textbf{2.23} & \textbf{0.29s} \\
\bottomrule
\end{tabular}
\caption{Performance evaluation of ET and other baselines on mCVRP30}
\label{tab:cvrp}
\end{table}

\end{document}